\documentclass[11pt]{article}

\usepackage[final]{acl}

\usepackage{times}
\usepackage{latexsym}

\usepackage{threeparttable}
\usepackage{multirow}
\usepackage{array}
\usepackage{pifont}
\usepackage{enumitem}
\usepackage[T1]{fontenc}

\usepackage[utf8]{inputenc}

\usepackage{microtype}

\usepackage{inconsolata}

\usepackage{graphicx}

\usepackage{listings}
\usepackage{booktabs}
\usepackage[table]{xcolor}
\usepackage{multirow}
\usepackage[most]{tcolorbox}
\usepackage{makecell}
\usepackage{colortbl}
\usepackage{pifont}
\usepackage{subfig}
\usepackage{float}
\usepackage{bbm}

\title{MedMCP-Calc: Benchmarking LLMs \\for Realistic Medical Calculator Scenarios via MCP Integration}

\newcommand*\samethanks[1][\value{footnote}]{\footnotemark[#1]}
\author{
 \textbf{Yakun Zhu\textsuperscript{1,2}}\thanks{~~Co-first authors}\space\space\space
 \textbf{Yutong Huang\textsuperscript{1}}\samethanks\space\space\space
 \textbf{Shengqian Qin\textsuperscript{1}}\samethanks\space\space\space
 \textbf{Zhongzhen Huang\textsuperscript{1}}\space\space\space
\\
 \textbf{Shaoting Zhang\textsuperscript{1}}\thanks{~~Corresponding author}\space\space\space
 \textbf{Xiaofan Zhang\textsuperscript{1,2}}\samethanks
\\
\\
 \textsuperscript{1}Shanghai Jiao Tong University,\space\space\space
 \textsuperscript{2}Shanghai Innovation Institute
\\
}

\begin{document}
\maketitle
\begin{abstract}
Medical calculators are fundamental to quantitative, evidence-based clinical practice. However, their real-world use is an adaptive, multi-stage process, requiring proactive EHR data acquisition, scenario-dependent calculator selection, and multi-step computation, whereas current benchmarks focus only on static single-step calculations with explicit instructions. To address these limitations, we introduce MedMCP-Calc, the first benchmark for evaluating LLMs in realistic medical calculator scenarios through Model Context Protocol (MCP) integration. MedMCP-Calc comprises 118 scenario tasks across 4 clinical domains, featuring fuzzy task descriptions mimicking natural queries, structured EHR database interaction, external reference retrieval, and process-level evaluation. 
Our evaluation of 23 leading models reveals critical limitations: even top performers like Claude Opus 4.5 exhibit substantial gaps, including difficulty selecting appropriate calculators for end-to-end workflows given fuzzy queries, poor performance in iterative SQL-based database interactions, and marked reluctance to leverage external tools for numerical computation. Performance also varies considerably across clinical domains.
Building on these findings, we develop CalcMate, a fine-tuned model incorporating scenario planning and tool augmentation, achieving state-of-the-art performance among open-source models\footnote{\url{https://github.com/SPIRAL-MED/MedMCP-Calc}}.
\end{abstract}

\section{Introduction}

\begin{figure}[t!]
  \includegraphics[width=\linewidth]{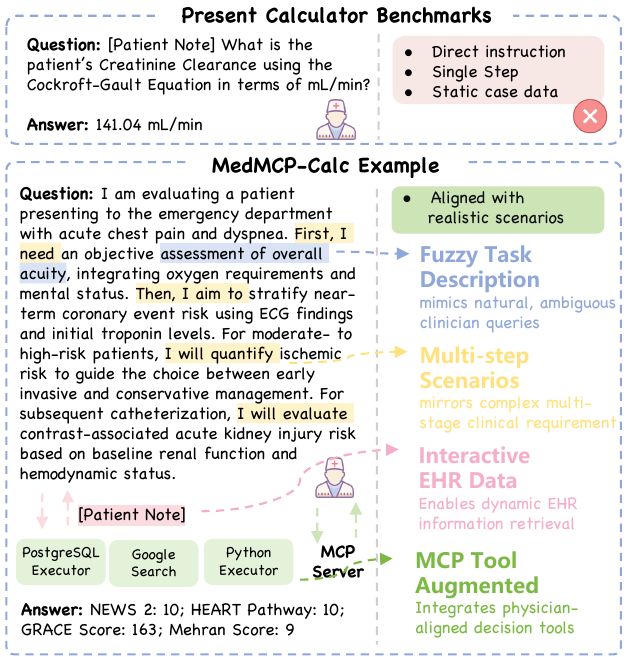}
  \caption{Existing benchmarks typically provide direct instructions with static case data for single-step computation. In contrast, MedMCP-Calc presents fuzzy task descriptions mimicking natural clinician queries, requires multi-step decision making across complex scenarios, enables dynamic interaction with structured EHR databases, and integrates MCP-based tools for physician assistants.}
  \label{fig-appetizer}
\end{figure}

Medical calculators are foundational to quantitative, evidence-based clinical practice. They transform patient-specific data into risk estimates, diagnostic thresholds, and treatment recommendations, thereby supporting diagnosis, triage, and therapeutic decision-making across diverse clinical settings. In practice, the use of medical calculators is rarely a single, isolated computation. Instead, it unfolds as an adaptive, multi-stage process embedded within broader clinical workflows. For example, in emergency chest pain triage, clinicians may first apply an initial stability screening (e.g., NEWS2), followed by risk stratification (e.g., HEART Pathway). These results subsequently inform downstream decisions, such as mortality prediction (e.g., GRACE) or procedure-related risk assessment (e.g., Mehran) prior to angiography. Throughout this process, clinicians must gather fragmented patient information from electronic health records (EHRs), determine which calculators are clinically appropriate as context evolves, and reconcile multiple quantitative outputs into actionable decisions.

Recent advances in large language models (LLMs) have sparked growing interest in the potential to support such complex, tool-intensive workflows~\cite{mialon2023gaia, patilberkeley, fan2024workflowllm, xie2024travelplanner, qiao2024benchmarking}. A key enabler is the Model Context Protocol (MCP)~\cite{anthropic2024mcp}, often described as the ``USB-C of AI,''~\cite{rick2025mcpusb}, which standardizes how LLM-based agents interact with heterogeneous external systems, including tools, APIs, databases, and other resources. By replacing bespoke, task-specific integrations with a uniform protocol, MCP enables a more extensible ecosystem in which agents can effectively operate with ``eyes and hands'' in real-world environments~\cite{hou2025model, yang2025survey, hasan2025model, wang2025mcp}.  In the medical domain, MCP is particularly promising for enabling ``clinical agentic workflows'', allowing LLMs to seamlessly interface with EHR databases and auxiliary tools such as search engines and medical calculators.

Despite the rapid adoption of MCP and the emerging body of MCP-based benchmarks, most existing applications focus on general-purpose domains rather than healthcare, and rarely address the stringent requirements of medical calculator use~\cite{wang2025mcp, luo2025mcp, wu2025mcpmark, liu2025mcpeval}. Current medical calculator benchmarks typically evaluate models under highly simplified conditions that deviate substantially from real-world clinical practice~\cite{khandekar2024medcalc, zhu2025menti, zhang2025cmedcalc}. Specifically, (1) test cases are often presented as clean, pre-packaged narratives processed in a single turn, rather than requiring iterative evidence gathering from noisy, structured EHR; (2) agents are explicitly instructed to invoke specific calculators, instead of responding to ambiguous, goal-driven queries that reflect natural clinical interactions; and (3) data and tool logic are tightly entangled, making it costly to integrate and maintain evolving calculators. These limitations create a substantial gap in assessing LLMs’ ability to perform realistic, multi-step clinical computations, hindering the development of trustworthy medical agents.

To address this gap, we introduce MedMCP-Calc, the first benchmark designed to evaluate LLMs on realistic medical calculator workflows integrated with MCP servers. MedMCP-Calc comprises 118 clinically grounded scenarios spanning four medical domains. As illustrated in Figure~\ref{fig-appetizer}, the benchmark incorporates three MCP-based environments to support realistic simulation: (1) a \texttt{PostgreSQL} server for iterative evidence acquisition from EHR databases, (2) \texttt{Google Search} for retrieving up-to-date references (e.g., calculator definitions and clinical guidelines), and (3) a \texttt{Python Executor} to enable robust and precise numerical computation. Task prompts in MedMCP-Calc are phrased as fuzzy, goal-oriented questions rather than explicit calculator instructions, better reflecting naturalistic clinician intent. The benchmark also offers an evaluation method for assessing calculator selection, evidence acquisition, and quantitative accuracy capabilities.

We conduct extensive experiments on 23 leading models. Across all models, including top-performing systems such as GPT-5 and Gemini-3-Pro-Preview, we observe substantial performance limitations, underscoring the inherent difficulty of solving tasks within realistic medical calculator workflows. Detailed analyses reveal several key challenges faced by current LLM agents. First, LLMs struggle to select all appropriate calculators required to complete an end-to-end workflow when confronted with fuzzy queries. Second, they exhibit poor performance in iterative database interactions involving SQL. Third, LLMs show a marked reluctance to leverage external tools for numerical computation. In addition, performance varies considerably across clinical domains, indicating the need for domain-specific optimization, especially for complex scoring systems and cognitive assessment scales. Collectively, these findings demonstrate the value of MedMCP-Calc in surfacing critical limitations and open challenges in realistic medical calculator workflows. Building on these, we develop \textbf{CalcMate}, a fine-tuned model incorporating scenario planning and aggressive tool augmentation, which achieves state-of-the-art performance among open-source models. 

Our contributions can be summarized as follows:
\begin{itemize}[noitemsep]
\item We introduce MedMCP-Calc, the first MCP benchmark for evaluating LLMs on realistic medical calculator workflow tasks.
\item We conduct comprehensive evaluations of 23 leading models, revealing critical limitations in real-world calculator task performance.
\item We propose CalcMate, demonstrating that scenario planning combined with tool-augmented training significantly improves performance on complex medical calculator tasks.
\end{itemize}

\section{Related Works}

\textbf{LLM Agents.} Modern LLMs~\cite{comanici2025gemini, QwenTeam2025Qwen3, team2025kimi, Anthropic2025Claude4} have evolved from instruction following to autonomous agents capable of planning and tool use, further enhanced by reinforcement learning~\cite{shao2024deepseekmath, putta2024agent, huang2025adaptive, huang2026real, huang2026does, zhang2026heterogeneous}. Early prompting strategies such as ReAct~\cite{yao2022react} and reflection~\cite{shinn2023reflexion} strengthened single-agent reasoning, while multi-agent frameworks like MetaGPT~\cite{hong2023metagpt} and AutoGen~\cite{wu2024autogen} coordinate complex workflows via explicit role assignment. Capabilities have expanded from API-based tool use~\cite{schick2023toolformer, qin2023toolllm, li2023api, qsq2025meta, liu2026rethinking} to multimodal interactions in digital environments~\cite{xie2024osworld, zheng2024gpt, wang2024mobile, Dong2025SemanticMA, Zhang2025DualAG}, including coding agents~\cite{yang2024swe} and GUI-based systems such as OpenAI's CUA~\cite{OpenAI2025ComputerAgent} and Anthropic's Computer Use~\cite{Anthropic2024ComputerUse}. This trend toward general-purpose agents motivates standardized protocols like the Model Context Protocol (MCP)~\cite{Anthropic2024ModelContext}. Yet such agents struggle with the precision demanded in medical settings~\cite{masayoshi2025ehr}, lacking designs that reliably couple structured EHR retrieval with accurate clinical computation, and often hallucinating in data extraction and calculation.

\textbf{MCP Benchmark.}
Recent works have introduced various benchmarks for evaluating agent performance within MCP systems. 
Early efforts like MCP-AgentBench~\cite{guo2025mcp} and MCPVerse~\cite{lei2025mcpverse} prioritized tool coverage and scalability, while MCP-RADAR~\cite{gao2025mcp} adapted static datasets such as HumanEval to MCP scenarios. 
However, these approaches often overlook the dynamic nature of real-world applications where ground truth evolves over time. 
To better capture task complexity, MCP-Bench~\cite{wang2025mcp} targets complex real-world tasks, 
and OSWorld-MCP~\cite{jia2025osworld} extends evaluation to operating system tool invocations. 
A methodological gap persists in evaluation metrics. While MCPEval~\cite{liu2025mcpeval} and LiveMCPBench~\cite{mo2025livemcpbench} adopt the ``LLM-as-a-Judge'' for scalability, they often suffer from evaluation biases and lack rigorous ground-truth validation for real-time outcomes. Recent advances address these limitations through high-fidelity execution environments. 
MCP-Universe~\cite{luo2025mcp} integrates authentic MCP servers with time-sensitive scenarios for long-horizon dynamic workflows
, while MCPMark~\cite{wu2025mcpmark} employs containerized settings with diverse CRUD operations and programmatic verification to ensure safety and reproducibility. 
Despite these advances, current MCP benchmarks remain largely domain-agnostic or focused on OS-level interactions, lacking frameworks tailored to vertical medical applications—encompassing long-horizon clinical decision-making, EHR database interactions, and authentic clinical tool use for literature retrieval and dosage calculation.

\begin{figure*}[ht!]
    \includegraphics[width=\linewidth]{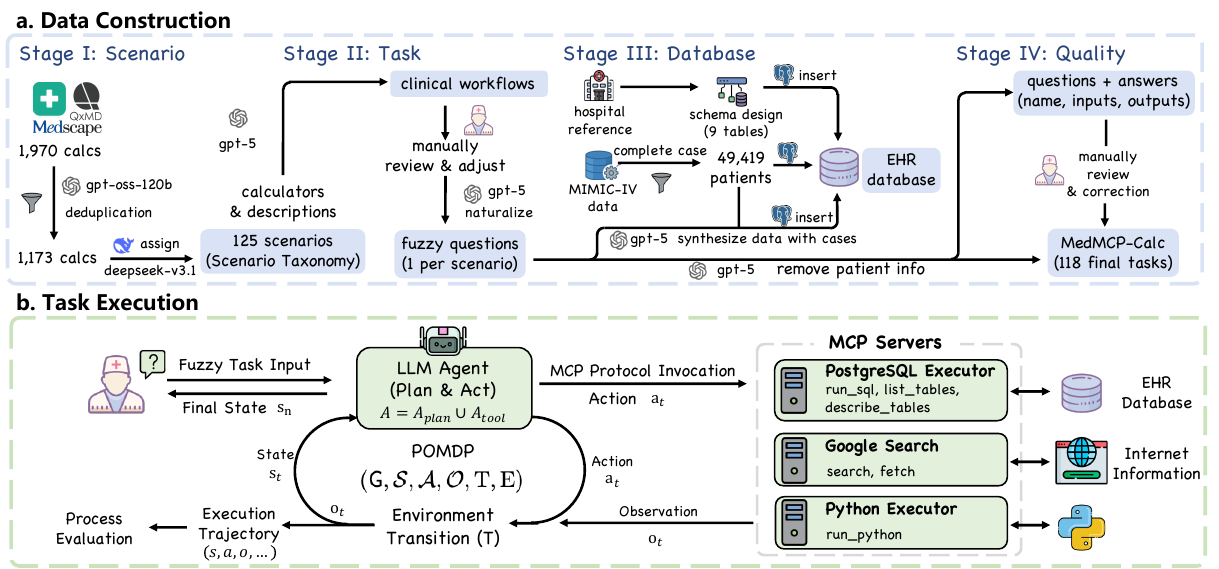}
    \caption{Overview of the MedMCP-Calc benchmark. (a) The data construction pipeline comprises four stages: Scenario Instantiation, Task Creation, Database Construction, and Quality Verification. (b) Task execution is formulated as a POMDP, where an LLM agent interacts with MCP servers to process clinical calculator tasks over realistic EHR databases.}
    \label{fig-main}
\end{figure*}

\textbf{Medical Calculators.}
LLM capabilities in quantitative reasoning~\cite{chen2025survey, chen2025code, zhang2023deep} have evolved from general mathematical problem-solving to specialized clinical calculations. Foundational benchmarks such as GSM8K~\cite{cobbe2021training} and MATH~\cite{hendrycks2021measuring} established the viability of chain-of-thought reasoning for structured logic~\cite{Dong2025InterleavedLV, Wang2025AutoregressiveSV, Li2025VideoProAP, wei2025time}, but translating this proficiency to medicine requires handling lengthy clinical contexts and strict adherence to domain-specific formulas. 
OpenMedCalc~\cite{goodell2023augmentation} first addressed this gap by demonstrating that augmenting LLMs with external calculation APIs significantly reduces errors. 
Building on this tool-use paradigm, AgentMD~\cite{jin2025agentmd} introduced an autonomous framework for curating and applying thousands of clinical calculators. 
MedCalc-Bench~\cite{khandekar2024medcalc} subsequently provided the first large-scale, human-verified evaluation dataset
, revealing LLM deficiencies in parameter extraction and formula selection. 
Recent works have further expanded this landscape: CalcQA~\cite{zhu2025menti} addresses an end-to-end clinical process involving nested tool calling within lengthy patient histories, while CMedCalc-Bench~\cite{zhang2025cmedcalc} extends assessment to Chinese medical environments. 
Despite these advances, existing studies predominantly formulate medical calculation as reading comprehension over static text, overlooking two critical aspects of real-world clinical workflows: (1) scenario task planning, requiring analysis of clinical scenarios and multi-step reasoning to orchestrate multiple calculators rather than executing isolated single-calculator instructions; and (2) proactive data acquisition, demanding dynamic interaction with clinical databases rather than passive extraction from pre-provided text.

\section{MedMCP-Calc Benchmark}
To rigorously evaluate LLMs' capabilities in solving complex, dynamic medical calculator tasks within realistic clinical scenarios via MCP servers, we introduce MedMCP-Calc.

\subsection{Formalization}
The task formulation and evaluation in MedMCP-Calc can be formally characterized as a Partially Observable Markov Decision Process (POMDP). Under multi-turn planning, execution, and observation, each task is represented as a POMDP tuple $(\mathsf{G}, \mathcal{S}, \mathcal{A}, \mathcal{O}, \mathrm{T}, \mathrm{E})$, where $\mathsf{G}$ denotes the task instruction and target outcome. For an LLM $m$ that executes a total of $n$ steps, the state space can be defined as $\mathcal{S} = \{s_0, s_1, \dots, s_n\}$ and the observation space as $\mathcal{O} = \{o_1, \dots, o_n\}$.

The action space $\mathcal{A}$ of the model $m$ encompasses both planning and tool invocation, formally expressed as $\mathcal{A} = \mathcal{A}_{\text{plan}} \cup \mathcal{A}_{\text{tool}}$. We adopt the MCP, where each tool action is defined as $\mathcal{A}_{\text{tool}} = \{(a_{\text{server\_name}}, a_{\text{tool\_name}}, a_{\text{args}}), \ldots\}$, specifying the MCP server, tool name, and tool parameters, respectively. At each round $t$, the agent determines the next action $a_t$ based on the current state $s_{t-1}$ and the preceding observation $o_{t-1}$. The execution of $a_t$ triggers a state transition and yields a new observation, which is formally captured by the transition function $\mathrm{T}: \mathcal{S} \times \mathcal{A} \rightarrow \mathcal{S} \times \mathcal{O}$. This iterative process continues until either the maximum number of rounds is exceeded or a final answer is produced.

For the final evaluation, given any LLM $\forall m \in \mathcal{M} = \{m_1, \ldots\}$ and agent framework $\forall f \in \mathcal{F} = \{f_1, \ldots\}$, we assess performance based on their trajectories. Specifically, we evaluate the accuracy of achieving task objectives through $\mathrm{E}_{\text{final}}: \mathsf{G} \times \mathcal{S} \rightarrow \{0, 1\}$. Beyond final outcomes, we also assess process accuracy, the correctness of intermediate steps during execution $\mathrm{E}_{\text{step}}: \mathcal{S} \times \mathcal{A} \times \mathcal{O} \rightarrow \{0, 1\}$. This involves rigorous validation of intermediate parameters by comparing the execution results of the tool calling against the ground-truth.

\subsection{Component}
MedMCP-Calc comprises four core components: a scenario taxonomy defining clinical task categories, a physician workstation simulating the clinical environment, naturalistic tasks reflecting real-world user behavior, and evaluation metrics for comprehensive assessment.

\textbf{Scenario Taxonomy.} We collaborated with physicians to develop a classification taxonomy encompassing four major clinical domains: \textit{Critical Care \& Perioperative Medicine},  \textit{Internal Medicine \& Organ Systems}, \textit{Neurology \& Psychiatry}, and  \textit{Special Populations \& Universal Tools}. This taxonomy identifies 125 distinct clinical use categories for medical calculators (e.g., ``Chest Pain and Acute Coronary Syndrome''), each involving multi-step, interdependent decision-making processes.

\textbf{Physician Workstation.} To replicate the clinical working environment, we designed a series of MCP servers: (1) a \texttt{PostgreSQL Executor} server providing \texttt{run\_sql}, \texttt{list\_tables}, and \texttt{describe\_tables} tools for EHR database interaction (we constructed an EHR database from MIMIC-IV data); (2) a \texttt{Google Search} server with \texttt{search} and \texttt{fetch} tools for retrieving up-to-date medical references; and (3) a \texttt{Python Executor} server with a \texttt{run\_python} tool for numerical calculations. This design mirrors how physicians retrieve patient data from hospital databases and consult medical resources during calculator use.

\textbf{Naturalistic Task.} Each task instance is anchored to a specific scenario and patient, accompanied by a fuzzy question that prompts contextual clinical reasoning rather than explicit calculator instructions. For example, instead of ``calculate the HEART score,'' the task asks for risk stratification based on clinical findings. Tasks encompass multiple procedural steps and branching decisions, requiring the LLM to iteratively select actions until reaching a final answer.

\textbf{Evaluation Metrics.} We evaluate both final outputs and intermediate processes: \textit{Task Fulfillment (TF)} measures the proportion of tasks that yield valid final outputs out of all evaluated tasks; \textit{Calculator Selection (CS)} measures the accuracy of calculator choices, computed as the average ratio of correctly selected calculators to the total number of calculators required per task; \textit{Quantitative Precision (QP)} measures the numerical accuracy of computed results, calculated as the average ratio of calculators with correct outputs to the total number of calculators invoked per task; and \textit{Evidence Acquisition (EA)} assesses the LLM's capability to proactively acquire relevant data from databases, computed as the average ratio of correctly extracted data items to the total number of required inputs across all calculators per task. Details are presented in Appendix~\ref{apx-exp-metrics}.

\begin{table*}[!t]
\centering
\resizebox{\linewidth}{!}{
\begin{tabular}{lccccccc}
\toprule

\textbf{Benchmark} & \makecell{\textbf{Interactive}\\\textbf{EHR Data}} & \makecell{\textbf{Fuzzy Task}\\\textbf{Description}} &
\makecell{\textbf{Multi-step}\\\textbf{Scenarios}} & 
\makecell{\textbf{Process}\\\textbf{Evaluation}} & \makecell{\textbf{Tool}\\\textbf{Augmented}} & \makecell{\textbf{MCP}\\\textbf{Ecosystem}} \\
\midrule

OpenMedCalc~\cite{goodell2023augmentation} & \ding{56} & \ding{56} & \ding{56} & \ding{56} & \ding{52} & \ding{56} \\
MedCalc-Bench~\cite{khandekar2024medcalc} & \ding{56} & \ding{56} & \ding{56} & \ding{56} &\ding{56} & \ding{56} \\
CalcQA~\cite{zhu2025menti} & \ding{56} & \ding{52} & \ding{56} & \ding{52} & \ding{52} & \ding{56} \\
CMedCalc-Bench~\cite{zhang2025cmedcalc} & \ding{56} & \ding{56} & \ding{56} & \ding{52} &\ding{56} & \ding{56} \\

\midrule

MedMCP-Calc & \ding{52} & \ding{52} & \ding{52} & \ding{52} & \ding{52} & \ding{52} \\
\bottomrule
\end{tabular}}
\caption{\textbf{Comparison with existing medical calculator benchmarks.} Existing benchmarks primarily evaluate models' abilities to memorize and apply static medical knowledge for calculator computations, making it difficult to assess their practical capabilities in real-world clinical scenarios that require proactive EHR data retrieval, user intent recognition, multi-step workflow resolution, and real-time tool integration. In contrast, MedMCP-Calc aligns with real-world complex calculator tasks and comprehensively evaluates the practical task-solving capabilities.}
\label{table-comparison}
\end{table*}

\subsection{Pipeline}
Based on the benchmark design, we present a detailed construction pipeline for generating tasks in the following parts. 

\textbf{Stage I: Scenario Instantiation.} Following the established scenario taxonomy, we instantiated specific scenarios by collecting medical calculators and mapping them to corresponding categories. We gathered all publicly available calculators from three widely-used medical platforms—MDCalc, Medscape, and QxMD—yielding 1,970 entries. To eliminate redundancy, we employed \texttt{GPT-OSS-120B} for two rounds of semantic similarity screening based on calculator names and descriptions, followed by manual removal of invalid entries, resulting in 1,173 unique calculators. Guided by the taxonomy, we then used \texttt{DeepSeek-V3.1} to assign applicable calculators to each of the 125 scenarios.

\textbf{Stage II: Task Creation.} For each scenario, we first established the corresponding clinical workflow. Specifically, we provided \texttt{GPT-5} with the assigned calculators and their functional descriptions, instructing it to construct coherent calculator sequences based on the scenario context. The generated workflows were then manually reviewed and adjusted to ensure alignment with real-world clinical practices, while the results demonstrated strong consistency. Subsequently, we used \texttt{GPT-5} to convert these workflows into naturalistic task instructions reflecting authentic user behavior. This process simulates real-world situations where users operate under incomplete information without explicit knowledge of procedural details, issuing underspecified instructions in a conversational tone and requiring the LLM to infer appropriate calculator sequences and execution strategies. Through this process, we constructed one task per scenario. Full prompt templates are provided in Appendix~\ref{apd:task_construction}.

\textbf{Stage III: Database Construction.} To enable realistic EHR interactions, we developed a database infrastructure under physician guidance (see Appendix~\ref{apd:recruitment}). We designed a relational schema modeled based on real hospital EHR systems, comprising nine tables (e.g., \texttt{patient\_information}), with detailed specifications provided in Appendix~\ref{apd:ehr}. We then extracted and filtered data from MIMIC-IV, mapping records to conform to our schema, and selected patients with complete records across all tables, yielding a cohort of 49,419 patients. To support task execution, we linked scenario tasks from Stage II with clinically similar existing patients, synthesized corresponding patient data with \texttt{GPT-5}, and validated the results~\cite{corbeil2025empowering}. Finally, we employed rule-based code generation to produce SQL statements for inserting task-specific records and removed patient information from task questions, thereby establishing the interactive platform for MedMCP-Calc. 

\textbf{Stage IV: Quality Verification.} We conducted multiple rounds of validation to ensure benchmark quality and enable rigorous evaluation. For generated task questions, we performed manual verification to ensure alignment with real-world clinical workflows and practical requirements. For task answers, calculator names, outputs, and input parameters, we referenced all elements against patient information through item-by-item review and correction, using the three aforementioned medical calculator websites as ground-truth references. After excluding data that did not meet quality standards (e.g., insufficient available calculators, inability for EHR data), we obtained a final set of 118 tasks.

\subsection{Characteristics}
Table~\ref{table-comparison} presents a comparative analysis between MedMCP-Calc and existing medical calculator benchmarks. MedMCP-Calc exhibits the following distinctive characteristics.

\textbf{Interactive EHR Data.} MedMCP-Calc incorporates a dynamic EHR database requiring active data retrieval, rather than providing static case information. This design enables realistic simulation of clinical data acquisition and evaluates models' capabilities in proactively identifying relevant patient information.

\textbf{Fuzzy Task Description.} Instead of explicit computational instructions, MedMCP-Calc employs naturalistic clinical queries that require models to interpret user intent and autonomously select appropriate calculators—reflecting how clinicians actually formulate requests in practice.

\textbf{Multi-step Scenarios.} MedMCP-Calc is constructed around comprehensive clinical workflows involving staged subtasks and logical reasoning chains, moving beyond single-calculator computations to evaluate authentic multi-step decision-making procedures.

\textbf{Process Evaluation.} Beyond final-answer accuracy, MedMCP-Calc assesses the information acquisition phase by analyzing database interactions and comparing extracted data against ground-truth calculator inputs—a dimension largely overlooked by existing benchmarks.

\textbf{MCP Tool Augmented.} Rather than relying on limited, calculator-specific tools, MedMCP-Calc adopts the MCP protocol with general-purpose servers for database interaction, web search, and computation. This approach enables access to over one thousand publicly available online calculators, supports real-time retrieval of current medical guidelines, and mirrors clinicians' actual workflows of information retrieval and computation.

\section{Experiments}

\begin{table*}[!ht]
\centering
\definecolor{headergray}{HTML}{EAEAEA}
\definecolor{rowgray}{HTML}{F7F7F7}
\setlength{\tabcolsep}{3pt}
\resizebox{\linewidth}{!}{
\begin{tabular}{l l ccc ccc ccc ccc cccc}
\toprule
& & \multicolumn{3}{c}{\textbf{\makecell{Crit. Care \\ \& Periop.}}}
& \multicolumn{3}{c}{\textbf{\makecell{Int. Med. \\ \& Organ Sys.}}}
& \multicolumn{3}{c}{\textbf{\makecell{Neuro. \\ \& Psych.}}}
& \multicolumn{3}{c}{\textbf{\makecell{Spec. Pop. \\ \& Univ. Tools}}}
& \multicolumn{4}{c}{\textbf{Overall}} \\

\cmidrule(lr){3-5} \cmidrule(lr){6-8}
\cmidrule(lr){9-11} \cmidrule(lr){12-14}
\cmidrule(lr){15-18}

\multirow{-2}{*}{\textbf{Model}} & \multirow{-2}{*}{\textbf{Param.}} 
& CS & EA & QP     
& CS & EA & QP        
& CS & EA & QP      
& CS & EA & QP    
& TF & CS & EA & QP   \\ 
\midrule

\rowcolor{headergray}
\multicolumn{18}{c}{\textit{\textbf{Proprietary LLMs}}} \\
\midrule

Claude Opus 4.5 & / & 71.30 & 73.51 & 36.00 & 69.63 & 73.89 & 37.35 & 60.02 & 65.83 & 30.37 & 56.77 & 63.41 & 23.70 & 100 & \textbf{66.66} & \underline{71.08} & \textbf{33.92} \\
Claude Sonnet 4.5 & / & 69.55 & 53.32 & 23.64 & 67.77 & 56.64 & 33.10 & 55.98 & 41.26 & 31.57 & 60.97 & 55.14 & 14.72 & 100 & 65.60 & 53.86 & \underline{27.80} \\
        
GPT-5 & / & 61.87 & 46.58 & 26.61 & 62.77 & 56.64 & 29.64 & 60.20 & 49.15 & 27.87 & 48.24 & 53.70 & 17.33 & 100.00 & 59.81 & 53.12 & 26.70 \\ 
Gemini-3-Pro & / & 57.59 & 58.68 & 24.70 & 56.37 & 59.10 & 31.17 & 48.42 & 47.07 & 21.92 & 46.71 & 46.49 & 18.10 & 97.46 & 54.05 & 55.45 & 26.49 \\ 
Gemini-2.5-Pro & / & 59.61 & 41.98 & 20.83 & 58.89 & 41.35 & 21.76 & 50.38 & 35.00 & 18.95 & 46.64 & 39.70 & 19.08 & 100.00 & 55.96 & 40.45 & 20.77 \\ 
\midrule

\rowcolor{headergray}
\multicolumn{18}{c}{\textit{\textbf{Open Source LLMs}}} \\
\midrule

GLM-4.7 &355 B& 50.86 & 46.41 & 11.74 & 51.03 & 52.14 & 17.57 & 49.46 & 51.19 & 17.10 & 57.43 & 47.72 & 13.49 & 96.61 & 51.89 & 50.06 & \underline{15.59} \\
Deepseek-V3.1 &671 B & 41.85 & 37.28 & 10.77 & 45.35 & 39.70 & 14.22 & 43.79 & 44.44 & 7.42 & 44.78 & 47.29 & 12.38 & 99.15 & 44.33 & 41.04 & 12.37  \\ 
Qwen3-235B-Instruct-2507 &235 B & 48.83 & 34.95 & 11.38 & 41.31 & 32.62 & 10.29 & 34.60 & 39.49 & 11.03 & 39.41 & 34.42 & 11.26 & 100.00 & 41.74 & 34.24 & 10.76  \\ 
Qwen3-235B-Thinking-2507 &235 B & 39.74 & 28.28 & 13.77 & 39.81 & 27.70 & 12.90 & 33.63 & 16.39 & 6.87 & 38.37 & 26.79 & 5.64 & 100.00 & 38.82 & 26.33 & 11.14  \\ 
Kimi-K2-Instruct &1 T & 40.54 & 26.73 & 7.50 & 43.52 & 28.62 & 9.07 & 38.00 & 33.55 & 11.08 & 41.85 & 24.68 & 4.92 & 96.61 & 41.95 & 28.14 & 8.27 \\
Qwen3-235B non-thinking &235 B & 32.34 & 41.55 & 8.17 & 32.40 & 38.50 & 7.22 & 26.84 & 33.83 & 7.50 & 32.43 & 36.55 & 7.11 & 100.00 & 31.73 & 38.26 & 7.44  \\ 
Qwen3-235B thinking &235 B & 30.94 & 29.64 & 5.60 & 34.67 & 31.41 & 9.61 & 32.25 & 17.58 & 6.07 & 33.71 & 39.17 & 7.19 & 100.00 & 33.42 & 30.60 & 7.91  \\ 
Qwen3-VL-235B &235B & 32.69 & 26.13 & 7.87 & 37.16 & 27.31 & 11.76 & 22.45 & 18.17 & 4.46 & 32.88 & 25.52 & 10.99 & 99.15 & 33.74 & 25.67 & 9.94  \\ 
Qwen3-Next-80B-Instruct &80 B & 39.89 & 18.19 & 7.27 & 42.62 & 24.57 & 9.99 & 32.57 & 7.39 & 2.86 & 29.20 & 27.27 & 3.83 & 99.15 & 38.57 & 21.64 & 7.53  \\ 
Qwen3-Next-80B-Thinking &80 B & 35.79 & 33.57 & 10.23 & 34.09 & 35.97 & 13.33 & 33.08 & 33.23 & 10.77 & 28.02 & 29.84 & 5.40 & 99.15 & 33.30 & 34.10 & 11.03  \\ 
Llama-3.3-70B &70 B & 8.04 & 24.56 & 2.40 & 11.57 & 23.49 & 2.61 & 15.03 & 12.85 & 4.46 & 13.75 & 24.08 & 2.89 & 100.00 & 11.60 & 22.55 & 2.83  \\ 
Llama-3.1-70B &70 B & 9.22 & 16.37 & 0.73 & 12.55 & 23.24 & 1.19 & 16.12 & 14.03 & 0.00 & 10.19 & 15.03 & 1.50 & 99.15 & 11.87 & 19.30 & 1.01  \\ 
Qwen2.5-72B &72 B & 29.93 & 41.66 & 8.04 & 33.21 & 38.81 & 8.45 & 17.75 & 33.06 & 4.14 & 25.46 & 37.38 & 5.83 & 97.46 & 29.37 & 38.49 & 7.41  \\ 
Qwen3-30B-Instruct-2507 &30 B & 29.99 & 23.53 & 6.07 & 32.41 & 24.74 & 5.94 & 18.44 & 17.25 & 2.86 & 24.59 & 26.77 & 7.98 & 100.00 & 28.91 & 23.94 & 5.95  \\ 
Qwen3-30B-Thinking-2507 &30 B & 34.21 & 18.92 & 6.93 & 29.87 & 22.95 & 10.08 & 22.46 & 13.25 & 4.37 & 28.87 & 12.84 & 3.40 & 100.00 & 29.74 & 19.23 & 7.60  \\ 
Qwen3-4B-Instruct-2507 & 4 B & 20.70 & 15.14 & 7.47 & 19.30 & 8.09 & 1.45 & 12.83 & 6.54 & 3.73 & 25.94 & 7.74 & 1.94 & 100.00 & 19.95 & 9.34 & 3.08 \\ 
\midrule

\rowcolor{headergray}
\multicolumn{18}{c}{\textit{\textbf{Medical LLMs}}} \\
\midrule

Baichuan-M2-32B & 32 B & 29.79 & 32.13 & 8.77 & 36.31 & 36.72 & 9.23 & 36.02 & 26.44 & 7.40 & 30.73 & 33.46 & 2.46 & 79.66 & 33.95 & 33.98 & 7.77  \\ 
MedGemma-27B-IT &27 B& 31.75 & 10.31 & 2.93 & 32.45 & 20.49 & 4.54 & 28.85 & 15.47 & 1.69 & 23.35 & 24.68 & 3.08 & 97.46 & 30.33 & 18.45 & 3.61 \\ 
\midrule

\rowcolor{headergray}
\multicolumn{18}{c}{\textit{\textbf{Ours}}} \\
\midrule
CalcMate-4B & 4 B & 73.90 & 78.19 & 20.14 & 54.00 & 69.74 & 14.05 & 49.08 & 76.87 & 11.76 & 49.94 & 61.93 & 13.70 & \textbf{100} & \underline{56.97} & \underline{71.06} & 15.02 \\
CalcMate-30B & 30 B & 73.60 & 73.99 & 20.95 & 62.76 & 76.14 & 20.73 & 60.97 & 75.48 & 17.95 & 67.57 & 74.38 & 18.46 & \textbf{100} & \textbf{65.66} & \textbf{75.31} & \textbf{20.07} \\

\bottomrule
\end{tabular}}
\caption{\textbf{Model Performance on MedMCP-Calc.} The evaluation encompassed model performance across four clinical domains (\textit{Critical Care \& Perioperative Medicine}, \textit{Internal Medicine \& Organ Systems}, \textit{Neurology \& Psychiatry}, and \textit{Special Populations \& Universal Tools}) and three evaluation metrics (Calculator Selection, Evidence Acquisition, and Quantitative Precision), along with the overall performance with Task Fulfillment. Best results are highlighted in \textbf{bold} (best) and \underline{underlined} (second-best) under two settings, all models and open-source models only.}
\label{tab:main_results}
\vspace{-5pt}
\end{table*}

\subsection{Settings}

To comprehensively evaluate the capabilities of existing models in our MedMCP-Calc, we benchmarked a diverse set of state-of-the-art models, encompassing proprietary LLMs, open-source LLMs, and domain-specific medical models. For proprietary models, we evaluated Claude Opus 4.5~\cite{anthropic2025claudeopus45}, Claude Sonnet 4.5~\cite{anthropic2025claudesonnet45}, GPT-5~\cite{openai2025gpt5}, Gemini-3-Pro-Preview~\cite{gemini3} and Gemini-2.5-Pro~\cite{comanici2025gemini}. For open-source models, we included GLM-4.7~\cite{5team2025glm45agenticreasoningcoding}, DeepSeek-V3.1~\cite{deepseekai2024deepseekv3technicalreport}, Kimi-K2-Instruct~\cite{team2025kimi}, Qwen3-Series~\cite{qwen3technicalreport}, Llama-3.3~\cite{dubey2024llama} and Llama-3.1~\cite{dubey2024llama}. For domain-specific models, we evaluated Baichuan-M2-32B~\cite{dou2025baichuan} and MedGemma-27B-IT~\cite{sellergren2025medgemma}. 

We adopted the widely recognized ReAct agent framework~\cite{yao2022react} for evaluation and employed DeepSeek-V3.1 for structured extraction of results. GPT-OSS~\cite{openai2025gptoss120bgptoss20bmodel} and MedGemma-4B-IT were excluded from this experiment due to poor instruction-following capabilities—GPT-OSS struggled to adhere to the ReAct framework, consistent with findings in MCP-Universe~\cite{luo2025mcp}, while MedGemma-4B-IT showed similar limitations likely attributable to its smaller model size. Additionally, we included our fine-tuned model, \textbf{CalcMate}, derived from Qwen3-4B-Instruct-2507 through specialized training on medical workflow scenarios with enhanced tool-use capabilities.

\subsection{Main Results}  \label{sec-main-result}

Table \ref{tab:main_results} presents the main evaluation results of LLMs on MedMCP-Calc. Based on these results, we draw the following conclusions. We present only the main findings here; further detailed analysis can be found in Appendix~\ref{apx-further-main}.

\noindent \textbf{Overall Observation.} 
Regarding the Task Fulfillment, experimental results confirm that mainstream models have achieved maturity in instruction-following capabilities, with near-perfect task fulfillment after minimal post-processing, demonstrating readiness for multi-turn ReAct-based interactions in complex clinical scenarios.

\textbf{Proprietary models outperform open-source counterparts.} 
Claude Opus 4.5 achieves the best overall performance, ranking first in both CS and QP and demonstrating exceptional capabilities across all metrics. Claude Sonnet 4.5 also shows competitive results with strong comprehensive task-handling ability. GPT-5 and Gemini-3-Pro follow closely but slightly lag behind. Among open-source models, GLM-4.7 delivers impressive results, closely trailing proprietary models on CS and EA while even surpassing Gemini-2.5-Pro on EA. DeepSeek-V3.1 also demonstrates robust performance.


\textbf{Models struggle with calculator selection in fuzzy questions.}
Although Claude Opus 4.5 (66.66) and Claude Sonnet 4.5 (65.6) achieve strong Calculator Selection scores alongside GPT-5 (59.81), most other models—including medical-specialized ones—cluster around 30. This reflects two key factors. First, calculators have been overlooked in certain model development, leaving specialized models like MedGemma with limited exposure to calculator-related content. Second, real-world calculator scenarios pose additional challenges: while Baichuan-M2 achieves respectable CS performance, its low TF score reveals difficulties with agent frameworks requiring tool interaction and long-context reasoning. Top-performing models demonstrate more consistent behavior, though with notable deviations on less common calculators.

\textbf{Models exhibit poor performance in iterative database interactions.}
Models are required to autonomously compose SQL queries to retrieve patient data, yet they exhibit poor planning capabilities and low robustness when handling complex task intents. To enhance SQL robustness, we incorporated \texttt{list\_tables} and \texttt{describe\_tables} tools into the experimental setup, as all models—particularly smaller ones—exhibit hallucination without explicit schema inspection: fabricating queries based on unfounded assumptions and persisting in erroneous approaches despite failure signals. This behavior may stem from the disproportionate representation of certain database schemas in training corpora, which biases models' default assumptions.

\textbf{LLMs show marked reluctance to leverage external tools for numerical computation.}
Success requires models to select the correct calculator, retrieve accurate data, comprehend computation rules, and perform numerical calculations. The suboptimal results stem from multiple factors. First, limited performance on CS and EA creates compounding effects. Second, models fail to reliably internalize calculator-specific rules; while high-performing models handle common calculators adequately, performance degrades substantially for less frequent ones and among other models. Third, models rarely invoke external tools or consult calculator references, instead relying predominantly on internal knowledge. Fourth, computational hallucinations occur across small and medium-sized models, with smaller models exhibiting frequent arithmetic errors. And despite these limitations, models show minimal inclination to leverage Python tools for computational assistance.

\textbf{Extended thinking improves computational precision but may impair tool utilization.}
All thinking models achieve higher QP than non-thinking counterparts, as deliberation enhances numerical calculation accuracy. However, thinking mode degrades EA (e.g., Qwen3-235B: 38.26 to 30.60), likely due to over-reliance on internal knowledge rather than external tools. Dedicated thinking models (e.g., Qwen3-Next-80B) show larger but less balanced gains compared to hybrid models.

\noindent \textbf{Domain Variability.} Performance varies across clinical domains, with all models achieving higher scores in \textit{Critical Care \& Perioperative Medicine} and \textit{Internal Medicine \& Organ Systems} than in \textit{Neurology \& Psychiatry} and \textit{Special Populations \& Universal Tools}. The latter domains involve complex scoring systems and cognitive scales that prove more challenging than standardized physiological formulas, with QP scores often falling below 10\% even for top-performing models.

\begin{table}[t]
\centering
\setlength{\tabcolsep}{3pt}
\definecolor{headergray}{HTML}{EAEAEA}
\definecolor{rowgray}{HTML}{F7F7F7}
\resizebox{0.95\linewidth}{!}{
\begin{tabular}{l cccc}
\toprule
\textbf{Model} & \textbf{SQL} & \textbf{Python} & \textbf{Search} & \textbf{Fetch} \\
\midrule
Claude Opus 4.5 & 5.19 & 3.93 & 0.08 & 0.08  \\
Claude Sonnet 4.5 & 5.16 & 1.68 & 0.1 & 0.08 \\

GPT-5 & 3.85 & 0.06 & 0.25 & 0.25  \\ 
Gemini-3-Pro & 4.41 & 0.65 & 0.92 & 0.14  \\ 
Gemini-2.5-Pro & 4.45 & 0.99 & 0.14 & 0.14 \\ 

GLM-4.7 & 4.31 & 0.82 & 0.18 & 0.2  \\ 
DeepSeek-V3.1 & 3.71 & 0.94 & 0.64 & 0.04  \\ 
Qwen3-235B-Thinking-2507 & 4.08 & 0.13 & 0.01 & 0  \\ 
Qwen3-235B-Instruct-2507 & 3.92 & 0.53 & 0 & 0  \\ 
Kimi-K2-Instruct & 2.61 & 0.7 & 0.09 & 0.02  \\ 
Qwen3-235B thinking & 2.94 & 0.22 & 0.2 & 0.2  \\ 
Qwen3-235B non-thinking & 3.16 & 0.84 & 0.06 & 0.02  \\ 
Qwen3-Next-80B-Thinking & 4.05 & 0.07 & 0.07 & 0.03  \\ 
Qwen3-Next-80B-Instruct & 5.18 & 0.39 & 0.1 & 0.01  \\ 
Llama-3.3-70B & 3.5 & 2.45 & 0.02 & 0  \\ 
Llama-3.1-70B & 3.42 & 3.41 & 0.08 & 0 \\ 

Qwen3-30B-Thinking-2507 & 2.81 & 0.18 & 0.02 & 0  \\ 
Qwen3-30B-Instruct-2507 & 3.81 & 0.93 & 0.07 & 0.03  \\ 
Qwen3-4B-Instruct-2507 & 1.9 & 0.13 & 0.02 & 0.01 \\

MedGemma-27B-IT & 3.54 & 0.27 & 0.19 & 0  \\ 
Baichuan-M2-32B & 4.46 & 0.24 & 0.03 & 0.04 \\ 

CalcMate-4B & 7.97 & 7.32 & 8.16 & 8.07  \\ 
CalcMate-30B & 8.34 & 8.21 & 8.49 & 8.42  \\ 
			
\bottomrule
\end{tabular}}
\caption{Comparison of average tool usage frequency and types across different models.}
\label{tab:tool_usage}
\vspace{-15pt}
\end{table}

\subsection{CalcMate}

Through experimental analysis, we identify two primary limitations affecting model performance: (1) unfamiliarity with clinical decision-making and planning, and (2) limited tool utilization within agent frameworks. To address these limitations, we investigate whether fine-tuning with scenario planning and tool augmentation can more effectively solve calculator tasks in clinical settings.

\noindent \textbf{Scenario Planning.} Experiments reveal that most models, except top-tier ones, fail to plan and execute tasks step by step when facing complex clinical workflows. This limitation likely stems from insufficient reasoning capabilities and limited exposure to complex clinical scenarios. To address this, we leverage existing calculators and use \texttt{GPT-5} to construct 1,000 scenario tasks and generate the corresponding global planning reasoning.

\noindent \textbf{Tool Augmentation.} Experiments reveal a pronounced tendency toward tool avoidance across all models. However, tool usage is crucial for mitigating knowledge and computational hallucinations, particularly for smaller models. Therefore, we adopt an aggressive tool augmentation strategy. Each calculator computation follows the ReAct format, incorporating complete calculator reference operations, database exploration and querying, and code-assisted computation. While \texttt{run\_sql} data is generated by \texttt{GPT-5}, all other steps are constructed using \texttt{Qwen3-235B-A22B-Instruct-2507-FP8}. 

\noindent \textbf{Evaluation Results.}
CalcMate achieves comprehensive improvements across all metrics. CS gains validate scenario planning for superior task decomposition; EA improvements (9.34 to 71.06) confirm the effectiveness of our tool augmentation strategy; QP gains demonstrate code-assisted tools' contribution. Notably, CalcMate achieves SOTA among open-source models and surpasses both source models used for training data construction, confirming that gains stem from methodological advances rather than data distillation alone.

\subsection{Tool Utilization} \label{sec-exp-tool}

To comprehensively analyze tool utilization strategies across models and examine the relationship between tool usage and task performance in clinical calculator scenarios, we investigated the usage patterns of primary functional tools (\texttt{run\_sql}, \texttt{run\_python}, \texttt{search}, and \texttt{fetch}) during task execution, as presented in Table~\ref{tab:tool_usage}. We present only the main findings here; further detailed analysis can be found in Appendix~\ref{apx-further-tool}.

Current models exhibit conservative tool utilization: while SQL invocation is relatively frequent (3–5 calls) due to data retrieval requirements, auxiliary tools are significantly underutilized—Python averages below 1 call for most models, with notable exceptions like Claude Opus 4.5 (3.93), and Search/Fetch usage approaches zero for several models (e.g., Qwen3-235B-Instruct-2507 records zero for both).

Tool usage correlates with specific performance metrics: SQL frequency positively correlates with EA (e.g., Gemini-3-Pro: SQL=4.41, EA=55.45), while Python usage associates with QP improvements (e.g., Gemini-2.5-Pro: Python=0.99, QP=20.77). Notably, thinking models reduce tool invocation without performance degradation, suggesting enhanced reasoning partially compensates for reduced tool usage.

Our tool augmentation strategy yields substantial gains. CalcMate exhibits dramatically higher utilization across all tools compared to baselines, which show pronounced tool avoidance. This tool engagement directly improves performance.

\section{Conclusion}
We introduce MedMCP-Calc, the first benchmark for evaluating LLMs on realistic medical calculator tasks through MCP tool integration, featuring fuzzy task descriptions, interactive EHR databases, and multi-step clinical scenarios across 118 tasks spanning 4 clinical domains.
Our evaluation of 23 models reveals critical limitations: even Claude Opus 4.5 achieves only 33.92\% calculation accuracy, with models exhibiting knowledge hallucinations, computational errors, and tool avoidance tendencies.
We develop CalcMate through scenario planning training and tool augmentation, achieving state-of-the-art performance and demonstrating that targeted training on clinical workflow planning effectively bridges the gap between current capabilities and real-world demands. We hope this work advances both clinical AI applications and model development.

\section*{Limitations}
Despite its rigorous design and careful consideration of clinical task characteristics, MedMCP-Calc has several limitations. First, while the benchmark references real hospital EHR systems to simulate authentic clinical environments, the current data association patterns may exhibit region-specific biases and cannot fully capture the heterogeneity of EHR systems across diverse global healthcare settings. Second, although CalcMate's ``proactive tool augmentation'' strategy yields substantial performance improvements, it significantly increases context length, resulting in elevated inference latency and computational overhead. Future work could explore optimization techniques such as knowledge distillation or long-context compression to enhance efficiency, particularly for deployment in resource-constrained clinical environments.

\section*{Ethical Considerations}
This study presents MedMCP-Calc, a benchmark for evaluating LLMs on medical calculator tasks in simulated clinical environments. Patient data were derived from MIMIC-IV, a publicly available, de-identified dataset compliant with HIPAA regulations and approved by an institutional review board. Synthesized records were generated to supplement task-specific scenarios, containing no real patient identifiers. We acknowledge that LLMs may exhibit knowledge hallucinations and computational errors in clinical calculations, as our experimental results demonstrate. The observed tool avoidance tendencies and calculation inaccuracies highlight the risks of deploying such systems without appropriate safeguards. MedMCP-Calc serves as an academic evaluation framework for assessing medical AI capabilities in calculator-based clinical workflows, not as a deployable clinical decision support system. Any practical application of models evaluated or trained on this benchmark requires rigorous clinical validation with human oversight. Decision-making authority must remain with qualified healthcare professionals, with model outputs serving only as assistive references subject to professional verification.

\newpage
\bibliography{custom}

\appendix
\clearpage

\section{Experiments Details}

\begin{table*}[ht]
\centering
\caption{Model information and abbreviations used in experiments.}
\setlength{\tabcolsep}{20pt}
\label{tab:model_info}
\resizebox{0.9\textwidth}{!}{%
\begin{tabular}{ll}
\toprule
\textbf{Model Full Name} & \textbf{Abbreviation} \\
\midrule
\multicolumn{2}{l}{\textit{Proprietary Models}} \\
\midrule
Claude Opus 4.5~\cite{anthropic2025claudeopus45} & Claude Opus 4.5 \\
Claude Sonnet 4.5~\cite{anthropic2025claudesonnet45} & Claude Sonnet 4.5 \\
GPT-5~\cite{openai2025gpt5} & GPT-5 \\
Gemini-3-Pro-Preview~\cite{gemini3} & Gemini-3-Pro \\
Gemini-2.5-Pro~\cite{comanici2025gemini} & Gemini-2.5-Pro \\
\midrule
\multicolumn{2}{l}{\textit{Open-Source Models}} \\
\midrule
GLM-4.7~\cite{5team2025glm45agenticreasoningcoding} & GLM-4.7 \\
DeepSeek-V3.1~\cite{deepseekai2024deepseekv3technicalreport} & DeepSeek-V3.1 \\
Kimi-K2-Instruct~\cite{team2025kimi} & Kimi-K2-Instruct \\
Qwen3-235B-A22B (Thinking)~\cite{qwen3technicalreport} & Qwen3-235B thinking \\
Qwen3-235B-A22B (Non-Thinking)~\cite{qwen3technicalreport} & Qwen3-235B non-thinking \\
Qwen3-235B-A22B-Thinking-2507~\cite{qwen3technicalreport} & Qwen3-235B-Thinking-2507 \\
Qwen3-235B-A22B-Instruct-2507~\cite{qwen3technicalreport} & Qwen3-235B-Instruct-2507 \\
Qwen3-VL-235B-A22B-Instruct~\cite{qwen3technicalreport} & Qwen3-VL-235B \\
Qwen3-Next-80B-A3B-Instruct~\cite{qwen3technicalreport} & Qwen3-Next-80B-Instruct \\
Qwen3-Next-80B-A3B-Thinking~\cite{qwen3technicalreport} & Qwen3-Next-80B-Thinking \\
Llama-3.3-70B-Instruct~\cite{dubey2024llama} & Llama-3.3-70B \\
Llama-3.1-70B-Instruct~\cite{dubey2024llama} & Llama-3.1-70B \\
Qwen2.5-72B-Instruct~\cite{qwen2.5} & Qwen2.5-72B \\
Qwen3-30B-A3B-Instruct-2507~\cite{qwen3technicalreport} & Qwen3-30B-Instruct-2507 \\
Qwen3-30B-A3B-Thinking-2507~\cite{qwen3technicalreport} & Qwen3-30B-Thinking-2507 \\
Qwen3-4B-Instruct-2507~\cite{qwen3technicalreport} & Qwen3-4B-Instruct-2507 \\
Baichuan-M2-32B~\cite{dou2025baichuan} & Baichuan-M2-32B \\
MedGemma-27B-IT~\cite{sellergren2025medgemma} & MedGemma-27B-IT \\
\bottomrule
\end{tabular}}
\end{table*}

\subsection{Models}

Table~\ref{tab:model_info} provides the full names, abbreviations, and references for all models evaluated in our experiments.

\noindent \textbf{Closed-source Models.} For proprietary models, we conducted all evaluations through their official APIs provided by the respective vendors. We used the default hyperparameters unless otherwise specified.

\noindent \textbf{Open-source Models.} For open-source models, we deployed all models using SGLang~\cite{zheng2024sglang} as the inference framework. The experiments were conducted on servers equipped with 1-32 NVIDIA H100 GPUs or 1-8 NVIDIA H200 GPUs, depending on the model size and memory requirements.

\subsection{Metrics} \label{apx-exp-metrics}
We evaluate both final outputs and intermediate processes through four complementary metrics:

\textbf{Task Fulfillment (TF)} measures the percentage of successfully completed quantitative assessments, reflecting the model's end-to-end capability to complete valid clinical calculation tasks.
$$\text{TF} = \frac{N_{\text{success}}}{N_{\text{total}}} \times 100\%$$
where $N_{\text{success}}$ denotes the number of tasks with valid final outputs, and $N_{\text{total}}$ represents the total number of evaluation tasks.

\textbf{Calculator Selection (CS)} measures the percentage of appropriate calculator choices, evaluating whether the model can correctly identify the required medical calculator based on clinical context.
$$\text{CS} = \frac{1}{N_{\text{total}}} \sum _{i=1}^{N_{\text{total}}} \frac{N^i_{\text{correct\_calc}}}{N^i_{\text{calc}}} \times 100\%$$
where $i$ represents the $i$-th task, $N^i_{\text{correct\_calc}}$ indicates the number of correctly selected calculators in the $i$-th task, and $N^i_{\text{calc}}$ represents the number of calculators required in the $i$-th task.

\textbf{Quantitative Precision (QP)} measures the numerical accuracy of computed results, assessing the model's ability to perform precise calculations.
$$\text{QP} = \frac{1}{N_{\text{total}}} \sum_{i=1}^{N_{\text{total}}} [\frac{1}{N^i_{\text{calc}}} \sum_{j=1}^{N^i_{\text{calc}}} \mathbbm{1}(|y_{i,j} - \hat{y}_{i,j}| \leq \epsilon)]$$
where $y_{i,j}$ and $\hat{y}_{i,j}$ denote the ground-truth and predicted values of the $j$-th calculator in the $i$-th task, respectively, and $\epsilon$ is the tolerance threshold for numerical equivalence.

\textbf{Evidence Acquisition (EA)} assesses the LLM's capability to proactively acquire relevant data from databases, measuring information retrieval completeness.
$$\text{EA} = \frac{1}{N_{\text{total}}} \sum_{i=1}^{N_{\text{total}}} [\frac{|E_i^{\text{extracted}} \cap E_i^{\text{gt}}|}{|E_i^{\text{gt}}|}] \times 100\%$$
where $E_i^{\text{gt}}$ represents the ground-truth set of required patient data inputs for task $i$, and $E_i^{\text{extracted}}$ denotes the set of data fields successfully extracted by the model.

\subsection{ReAct Configuration} \label{apd:react}

All models are evaluated under the ReAct~\cite{yao2022react} agent framework. We adopt ReAct as it is the standard framework for MCP benchmarks, widely used in recent work such as MCP-Universe~\cite{luo2025mcp} and MCP-Bench~\cite{wang2025mcp}, which ensures direct comparability with the broader MCP community and isolates model capability from the confounds of framework design.

\textbf{Setup.} The maximum number of interaction rounds is set to 80. Each round consists of three steps: (1) \textbf{Thought}---the model analyzes the current state and reasons about the next action; (2) \textbf{Action}---the model issues an MCP tool call in JSON format with fields \texttt{server}, \texttt{tool}, and \texttt{arguments}; (3) \textbf{Observation}---the model receives the structured tool output and incorporates it into subsequent reasoning.

\textbf{ReAct Prompt:}
\begin{lstlisting}[breaklines=true, breakatwhitespace=true, basicstyle=\small\ttfamily, columns=fullflexible]
You are an expert Clinical Assessment Assistant and ReAct (Reasoning and Acting) agent.

Your primary role is to perform accurate medical calculations and clinical assessments based on patient-specific data retrieved from the EHR database.

You need to answer the following question:
Question: {{QUESTION}}

Your goal is to reason about the question and decide on the best course of action to answer it accurately.
You need to choose the appropriate tool based on the question. If no tool is needed, reply directly.
Please use only the tools that are explicitly defined below.

Here are the tools you have:
{{TOOLS_PROMPT}}


{% if HISTORY is defined and HISTORY|length %}
Previous steps and results:

{{HISTORY}}
{% else %}
Previous steps and results: EMPTY
{% endif %}

Instructions:
1. Analyze the query, previous reasoning steps, and results.
2. Decide on the next action: use a tool or provide a final answer.
3. Your MUST output the final answer within {{MAX_STEPS}} steps.
4. Respond in the following JSON format:

If you need to use a tool:
{
    "thought": "Your detailed reasoning about what to do next",
    "action": {
        "reason": "Explanation of why you chose this tool",
        "server": "server-name",
        "tool": "tool-name",
        "arguments": {
            "argument-name": "argument-value"
        }
    }
}

If you have enough information to answer the query:
{
    "thought": "Your final reasoning process to derive the answer.",
    "answer": "Final answer to the query"
}

Remember:
- Be thorough in your reasoning.
- Use tools when you need more information.
- Always base your reasoning on the actual results from tool use.
- If a tool returns no results or fails, acknowledge this and consider using a different tool or approach.
- Provide a final answer in the JSON when you're confident you have sufficient information.
- The response must be in a valid JSON format. Only JSON output is required, with no additional content.
\end{lstlisting}

\subsection{CalcMate Training} \label{apd:calcmate_training}

\textbf{Training Data.} The training set consists of 1{,}005 scenario tasks with a total of 9{,}097 calculator invocations (an average of 9.05 per task), spanning 637 unique calculators---54.31\% coverage of the full 1{,}173-calculator pool used in MedMCP-Calc. Tasks are distributed across the four clinical domains as \textit{Critical Care \& Perioperative Medicine} (440), \textit{Internal Medicine \& Organ Systems} (345), \textit{Neurology \& Psychiatry} (66), and \textit{Special Populations \& Universal Tools} (154), providing balanced coverage of clinical workflows. Within each trajectory, \texttt{GPT-5} is used only to produce the global planning trace (approximately 10\% of the data), while the remaining agent-execution content---\texttt{search}/\texttt{fetch} for calculator references, \texttt{describe\_tables} and SQL queries for database exploration, and \texttt{run\_python} for computation---is generated by \texttt{Qwen3-235B-A22B-Instruct-2507-FP8}. The training signal therefore primarily teaches systematic tool-use behavior rather than distilling GPT-5's medical knowledge, which is consistent with our observation that CalcMate surpasses GPT-5 by invoking external tools far more aggressively (cf.\ Python $8.21$ vs.\ $0.06$, Search $8.49$ vs.\ $0.25$ in Table~\ref{tab:tool_usage}) instead of relying on parametric knowledge.

\textbf{Training Setup.} CalcMate is obtained via full-parameter supervised fine-tuning on \texttt{Qwen3-4B-Instruct-2507} (CalcMate-4B) and \texttt{Qwen3-30B-A3B-Instruct-2507} (CalcMate-30B) using the LLaMA-Factory framework. We train for 1 epoch with a learning rate of $1\times10^{-5}$, a cosine learning-rate scheduler, and a warmup ratio of 0.1. The per-device batch size is 1 with 8 gradient accumulation steps. To accommodate the long multi-turn tool-use trajectories produced by our tool-augmentation strategy, we set the maximum sequence length to 150{,}000 tokens.

\textbf{System and Hardware.} Training uses bf16 mixed precision with FlashAttention-2 and gradient checkpointing for memory efficiency. Distributed training is conducted via DeepSpeed ZeRO Stage~3 with sequence parallelism (size~=~8). CalcMate-4B is trained on $8\times$ NVIDIA H100 GPUs, and CalcMate-30B on $32\times$ NVIDIA H100 GPUs.

\section{Further Analysis}

\subsection{Main Results}  \label{apx-further-main}

Table~\ref{tab:main_results} presents the main evaluation results of LLMs on MedMCP-Calc. In Section~\ref{sec-main-result}, we provide an analysis of the principal findings. In this appendix, we further extend our analysis through a detailed examination of more fine-grained model behaviors.

\noindent \textbf{Overall Observation.} 
Regarding the Task Fulfillment, experimental results indicate that mainstream models have achieved maturity in instruction-following capabilities. Although certain models (e.g., GPT-OSS) do not adhere to the ReAct framework, and others (e.g., Gemini-2.5-Pro) occasionally exhibit JSON format deviations, the vast majority achieve near-perfect task fulfillment after regular expression cleaning. This demonstrates that current models possess the fundamental capability for multi-turn interactions following the ReAct framework or tool-use protocols in complex clinical scenarios.

Regarding reasoning modes, extended thinking generally enhances computational precision; however, for agentic tasks and tool utilization, thinking models do not necessarily yield positive gains. Our experiments include the hybrid thinking model Qwen3-235B operating in different modes, along with three dedicated thinking models: Qwen3-235B-2507, Qwen3-Next-80B, and Qwen3-30B-2507. Results reveal that on Quantitative Precision, thinking models demonstrate a consistent advantage—all models achieve higher QP than their non-thinking counterparts, suggesting that extended deliberation enables more accurate handling of complex medical formula transformations and numerical calculations. Despite these improvements, all thinking models exhibit degradation in either CS or EA. For instance, the EA of Qwen3-235B decreases from 38.26 to 30.60 when the thinking mode is enabled, possibly reflecting a tendency to over-rely on internal knowledge during deep reasoning, thereby weakening external tool interaction capabilities. Furthermore, performance variance among hybrid thinking models is smaller than that observed in dedicated thinking models. When switching to thinking mode, Qwen3-235B exhibits more balanced yet conservative gains, with modest improvements in both Calculator Selection and Quantitative Precision. In contrast, dedicated thinking models such as Qwen3-Next-80B achieve more substantial improvements in both QP and EA.

\noindent \textbf{Domain Variability.} Performance varies notably across the four clinical domains. All models perform better in \textit{Critical Care \& Perioperative Medicine} and \textit{Internal Medicine \& Organ Systems} than in \textit{Neurology \& Psychiatry} and \textit{Special Populations \& Universal Tools}. For instance, models such as GPT-5 and Gemini-3-Pro maintain relatively high CS and EA in the former two domains, likely due to the standardized nature of physiological parameters and diagnostic protocols. In contrast, \textit{Neurology \& Psychiatry} poses a notable challenge, particularly in QP. Even high-performing models such as Deepseek-V3.1 and GLM-4.7 show sharp QP declines in this domain, often falling below 10\%. This suggests that clinical calculations involving complex scoring systems and cognitive scales are more difficult for LLMs to execute accurately than direct physiological formulas.

\noindent \textbf{Our Model.}
CalcMate achieves comprehensive improvements across all metrics. Even at the 4B parameter scale, it substantially outperforms the base model Qwen3-4B-Instruct-2507 and approaching state-of-the-art performance among open-source models. CalcMate-30B further establishes a clear lead over open-source alternatives, surpassing or approaching Gemini-2.5-Pro.

First, CalcMate-4B surpasses all models except GPT-5 in CS, while CalcMate-30B even exceeds GPT-5, demonstrating the effectiveness of scenario planning training and indicating superior task planning capabilities for calculator utilization. Second, EA improves substantially from 9.34 to 71.06, with the model invoking the \texttt{run\_sql} tool more than twice as frequently as other models, confirming that our aggressive tool augmentation strategy significantly enhances database interaction capability. Third, QP improves from 3.08 to 15.02—approaching GLM-4.7's 15.59—demonstrating the contribution of code-assisted tools. The relatively modest QP gains compared to CS and EA may reflect the inherent limitations of a 4B-parameter model in mathematical computation and coding; while code assistance improves calculation outcomes, these gains remain constrained by the model's fundamental capacity. The CalcMate-30B achieves substantially higher results (20.07), approaching proprietary model performance and validating that our method yields significant improvements in quantitative precision.

Notably, although our training data is constructed partially from GPT-5 and predominantly from Qwen3-235B-A22B-Instruct-2507-FP8, CalcMate surpasses both source models in performance, demonstrating that our improvements stem from methodological advances rather than inherited capabilities. Overall, these results validate our training strategy combining scenario planning with tool augmentation.

\subsection{Tool Utilization} \label{apx-further-tool}

Table~\ref{tab:tool_usage} presents tool usage patterns across different models. Section~\ref{sec-exp-tool} analyzes the principal findings, while this appendix extends the analysis to more fine-grained model behaviors.

Current models generally exhibit a conservative tendency toward tool underutilization. While all models demonstrate relatively high SQL invocation frequency (averaging 3–5 calls)—driven by the inherent task requirement of retrieving patient data from databases—they significantly underutilize auxiliary tools. For instance, most models invoke the Python tool fewer than once on average, with Qwen3-Next-80B-Thinking and GPT-5 averaging merely 0.07 and 0.06 calls, respectively. Search and Fetch tool usage is even sparser: Qwen3-235B-Instruct-2507 records zero invocations for both, and Qwen3-VL-235B shows zero Fetch calls. Even DeepSeek-V3.1, which exhibits relatively superior performance, achieves only 0.64 Search invocations. These findings suggest that current models tend to rely on parametric knowledge rather than proactively leveraging external tools for information retrieval or numerical computation, reflecting conservative tool utilization strategies.

Model performance exhibits certain correlations with tool usage strategies, while thinking models reduce tool invocation without sacrificing performance. First, SQL invocation frequency demonstrates a positive correlation with the EA metric. High-EA models such as GPT-5 (SQL=3.85, EA=53.12) and Gemini-3-Pro (SQL=4.41, EA=55.45) maintain relatively high SQL invocation frequencies, whereas Llama-3.3-70B (SQL=3.50, EA=22.55) and Qwen3-4B-Instruct-2507 (SQL=1.90, EA=9.34) exhibit lower EA corresponding to reduced SQL usage. This pattern is intuitive, as SQL serves as the primary mechanism for data retrieval and database interaction. Second, Python invocation frequency is associated with the QP metric. Gemini-2.5-Pro (Python=0.99, QP=20.77) and Qwen2.5-72B (Python=1.85, QP=7.41) suggest that Python invocation contributes to improved computational accuracy. And, most models exhibit relatively low reliance on Search and Fetch tools. Notably, comparing thinking models with their non-thinking counterparts reveals that the former generally demonstrate reduced frequencies of Python and external tool invocations without significant performance degradation. This suggests that enhanced internal reasoning capabilities can partially compensate for reduced tool usage—particularly for mathematical computations, where significantly reduced Python invocation does not compromise calculation performance.

Our tool augmentation strategy yields substantial performance gains. CalcMate-30B exhibits significantly higher tool utilization across all categories—SQL (8.34), Python (8.21), Search (8.49), and Fetch (8.42)—compared to baseline models, which show pronounced tool avoidance beyond basic SQL queries (e.g., Qwen3-235B-Instruct-2507: SQL 3.92, Python 0.53, Search 0, Fetch 0). This comprehensive tool engagement translates directly into performance improvements. Elevated SQL usage correlates with CalcMate's superior EA scores (75.31 vs. 34.24 for base Qwen3-235B), as thorough database exploration enables accurate clinical data extraction. Similarly, intensive Python utilization (8.21 vs. 0.53) corresponds to substantial QP gains (20.07 vs. 10.76), confirming that code-assisted computation effectively mitigates calculation errors. The dramatic increase in Search and Fetch operations (8.49/8.42 vs. near-zero for most baselines) demonstrates that our complete calculator reference pipeline substantially addresses knowledge hallucination—a critical limitation of smaller models. Therefore, even with only 4B parameters, CalcMate-4B achieves state-of-the-art performance among open-source models by leveraging tool assistance to reduce both computational and knowledge hallucinations, validating that systematic tool augmentation can compensate for limited model capacity in clinical calculator tasks.

\textbf{Efficiency of Tool Engagement.} A natural follow-up question is whether CalcMate's elevated tool-invocation counts reflect efficient behavior or wasteful looping on the same turns. Inspecting trajectories, we find that the base Qwen3-4B-Instruct-2507 frequently exhibits a trial-and-error pattern: it issues repeated near-duplicate SQL queries because of schema hallucination, guesses field names instead of calling \texttt{describe\_tables}, performs shallow \texttt{search} without subsequent \texttt{fetch}, and tries to self-correct arithmetic errors through more internal reasoning rather than invoking \texttt{run\_python}. These loops inflate round counts without producing useful evidence and often fail to converge. After tool-augmented training, CalcMate instead follows a structured execution pattern: when facing an unfamiliar calculator it first \texttt{search}es and then \texttt{fetch}es to acquire authoritative reference knowledge; it then calls \texttt{describe\_tables} to confirm schema before issuing targeted SQL; and for complex arithmetic,, it systematically routes to \texttt{run\_python} rather than relying on internal computation. The joint gains in EA and QP (Table~\ref{tab:main_results}) confirm that the increased tool frequency corresponds to purposeful, coordinated tool use rather than redundant looping, indicating that our augmentation teaches how to use tools rather than merely encouraging more calls.

\subsection{Extra Calculators Found Analysis.}

\begin{figure}[ht!]
    \includegraphics[width=\linewidth]{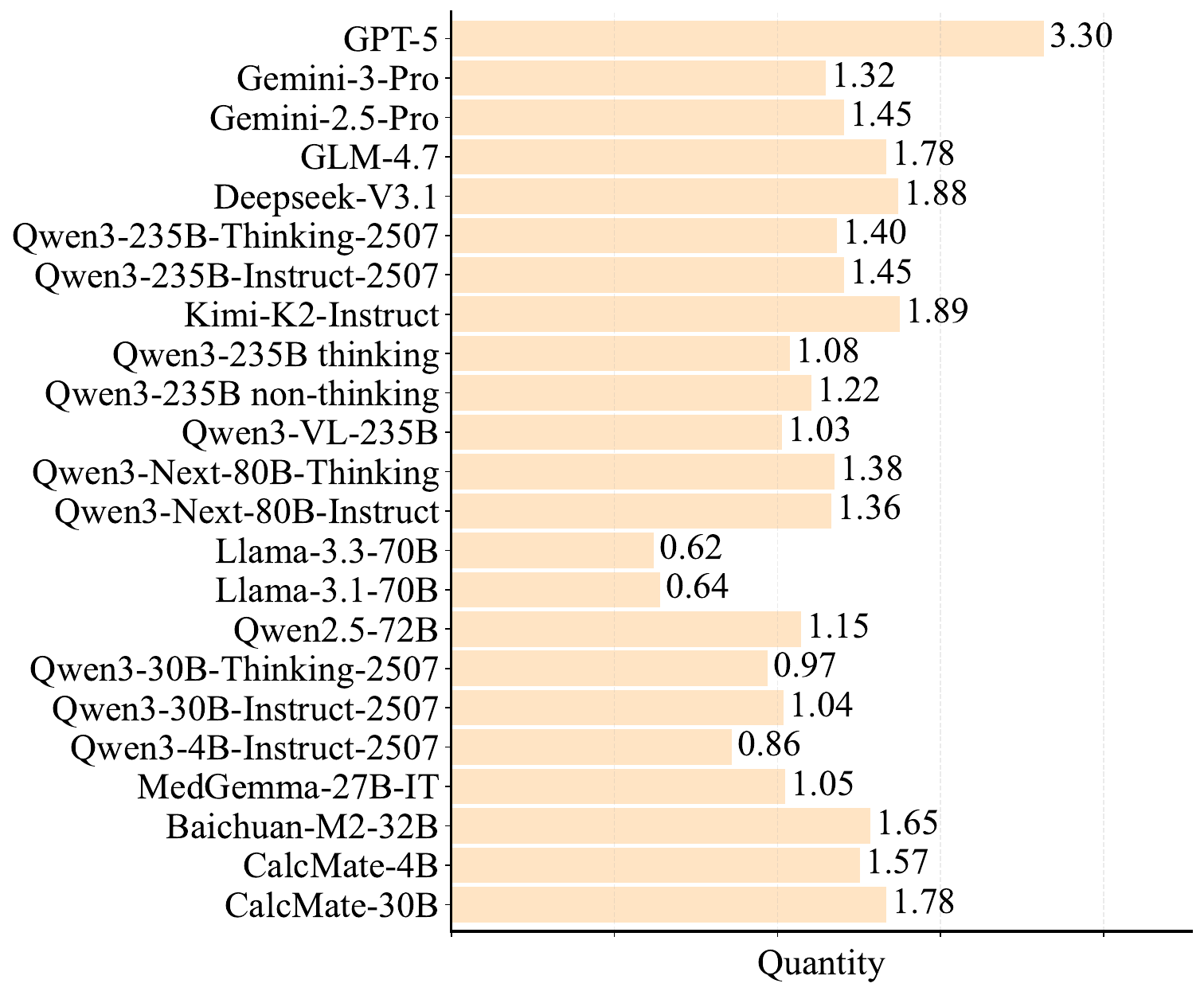}
    \caption{\textbf{Extra Calculators Found.} }
    \label{fig:extra_calc_found}
\end{figure}

During evaluation, we measure the average number of times a model selects calculators other than the ground truth calculator per task, as shown in Figure~\ref{fig:extra_calc_found}.

\textbf{Overall distribution.}
For most models, the number of extra calculator invocations falls within a relatively narrow and stable range. On average, models trigger approximately one to two extra calculator calls per task, with only minor differences across models. For example, core models from the Gemini, Qwen, and DeepSeek families mostly exhibit extra invocation counts in the range of 1.0–1.8, indicating a restrained and balanced level of exploration when using calculator tools.

\textbf{Outlier models and potential causes.}
GPT-5 exhibits an extra calculator invocation count of 3.3, making it the only model exceeding three calls and placing it significantly above all other models. In contrast, Llama-3.3-70B and Llama-3.1-70B show markedly lower values of 0.62 and 0.64, respectively, forming the lowest-performing cluster in this evaluation.

Given GPT-5’s strong overall performance among proprietary models, its substantially higher extra invocation rate likely reflects a more proactive strategy of exploring multiple calculator constructions or computation paths during task execution. This behavior may help cover diverse task requirements and increase the likelihood of selecting the correct calculator. Although such an aggressive strategy may involve some degree of redundant or unnecessary attempts, GPT-5 still achieves competitive performance on CS, QP, and EA, suggesting that increased extra invocations do not come at the cost of computational accuracy or data extraction quality.

The Llama-3 series ranks near the bottom on both CS and QP. Its lower frequency of extra calculator invocations may indicate a stronger reliance on internal parametric knowledge or fixed reasoning patterns, with fewer explicit selections of medical calculators. Combined with the relatively high frequency of Python tool usage observed during evaluation, this suggests that Llama models tend to convert parts of the computation directly into code execution rather than explicitly constructing or switching among multiple calculator logics.

\subsection{Failure Pattern Analysis} \label{apx-failure-analysis}

To understand where in the pipeline errors occur, we present a detailed failure analysis of a representative case — an ICU critical care scenario involving ARDS diagnosis, ventilator parameter optimization, pulmonary function assessment, and ECMO decision-making. In this case, GPT-5 successfully computed 2 of 10 ground-truth calculators (A-a O\textsubscript{2} Gradient = 434.9 and P/F Ratio = 85.0). The remaining failures fall into three categories:

Calculator Omission (4 calculators, $\sim$52.5\% of errors).
Despite being explicitly mentioned in the workflow, GPT-5 directly ignored four calculators: the Arterial Blood Gas Analyzer, Berlin Criteria for ARDS, Recruitment to Inflation Ratio, and PRESET Score. The first two are commonly used in ICU and emergency settings and are critical for ARDS scenarios. The omission likely stems from its limited five-step decision process, which lacks comprehensive planning and full workflow awareness. The latter two are relatively specialized, suggesting gaps in the model's domain-specific knowledge. Overall, the primary causes are the lack of global planning and knowledge omission.

Correct Calculator, Incorrect Result (3 calculators). This affected SpO\textsubscript{2}/FiO\textsubscript{2} Ratio, Static Lung Compliance, and Murray Score. We identified three underlying causes: (1)~the model simply forgot to perform the calculation despite strong context-handling capabilities; (2)~most frequently, GPT-5 failed to retrieve the necessary values from the database — the model tends to query data for multiple calculators simultaneously, leading to missed values, and upon encountering missing data, it hallucinated that the data was unavailable rather than re-querying; (3)~arithmetic errors, which were relatively rare for GPT-5 but occur frequently in smaller models. Additionally, during database interactions, the model often guessed field names instead of first performing exploratory queries — another form of overconfident hallucination. In summary, forgetting and hallucination are the main drivers of these failures.

Wrong Calculator Selection (1 calculator).
The expected calculator was ETT Depth, but the model selected Tidal Volume instead. These two serve fundamentally different purposes; the model likely chose Tidal Volume due to lexical similarity while overlooking the key term ``Depth.'' An incorrect ETT depth can result in single-lung ventilation or vocal cord injury, making this a clinically serious error.

Calculator omission accounts for approximately 52.5\% of errors, while incorrect results and wrong selection together account for approximately 47.2\%, with the remaining 0.3\% attributed to other errors. How to enable LLM foundation models to better learn agentic workflow reasoning and action remains an important open question.

\subsection{End-to-End Data Example} \label{apx-e2e-example}

To illustrate how our benchmark evaluates complete model behavior---spanning fuzzy natural-language input, multi-step SQL retrieval, calculator selection, arithmetic, and final interpretation---we present one full data point corresponding to the failure analysis case in Appendix~\ref{apx-failure-analysis}. This task requires a 10-step clinical workflow for ARDS assessment and ventilator optimization, drawn from Critical Care \& Perioperative Medicine.

\textbf{Fuzzy Query.}
The task is presented as a natural-language physician query, deliberately omitting explicit calculator names. A condensed version is shown below; the full text is included in our released dataset:

\begin{lstlisting}[breaklines=true, breakatwhitespace=true, basicstyle=\small\ttfamily, columns=fullflexible]
I'm seeing an adult in the ICU with diffuse pneumonia who's sliding into acute hypoxemic respiratory failure, and I'm worried this could be evolving ARDS. Oxygen needs have been climbing over the past several hours - I want a quick, bedside sense of how bad the oxygenation really is just from the pulse ox and the current oxygen setting. Once we have the [arterial blood] gas, I need to understand whether this is mostly shunt or V/Q mismatch by looking at the alveolar-arterial oxygen difference. [Confirm ARDS,] intubate and run a lung-protective strategy... look at static compliance... formal bedside check of recruitability... if severe hypoxemia persists, open the ECMO conversation. I need proper calculations on this.
\end{lstlisting}

This query requires the model to: (1)~identify which calculators are needed, (2)~retrieve patient data via SQL from an EHR-structured database, (3)~execute each calculation, and (4)~make data-driven clinical recommendations.

\textbf{Patient Data.}
The benchmark provides an EHR database for a de-identified 58-year-old male patient (P006\_efad2a48) admitted to the MICU for severe hypoxemic respiratory failure due to bilateral pneumonia (ICD-10: J80, J18.9). Key parameters stored across nine tables are summarized in Table~\ref{tab:e2e-patient}.

\begin{table}[h]
\centering
\small
\begin{tabular}{lll}
\toprule
\textbf{Source} & \textbf{Parameter} & \textbf{Value} \\
\midrule
\texttt{vital\_signs} & Height & 178\,cm (70\,in) \\
 & SpO\textsubscript{2} & 91\% \\
 & MAP & 75\,mmHg \\
\midrule
\texttt{laboratory\_result} & ABG: pH & 7.31 \\
 & ABG: PaCO\textsubscript{2} & 54\,mmHg \\
 & ABG: PaO\textsubscript{2} & 68\,mmHg \\
 & ABG: HCO\textsubscript{3}\textsuperscript{--} & 26\,mEq/L \\
 & ABG: FiO\textsubscript{2} & 80\% \\
 & Lactate & 3.8\,mmol/L \\
 & \ldots & \ldots \\
\bottomrule
\end{tabular}
\caption{Key patient parameters for the end-to-end example (subset).}
\label{tab:e2e-patient}
\end{table}

\textbf{Ground-Truth Annotation.}
The ground-truth defines a 10-step conditional workflow. Table~\ref{tab:e2e-gt} shows a subset of the steps.

\begin{table}[h]
\centering
\small
\resizebox{\columnwidth}{!}{
\begin{tabular}{clll}
\toprule
\textbf{Step} & \textbf{Calculator} & \textbf{Key Inputs} & \textbf{Expected} \\
\midrule
1 & SpO\textsubscript{2}/FiO\textsubscript{2} Ratio & SpO\textsubscript{2}=91\%, FiO\textsubscript{2}=80\% & 113.8 \\
2 & ABG Analyzer & pH=7.31, PaCO\textsubscript{2}=54, HCO\textsubscript{3}\textsuperscript{--}=26 & Resp.\ Acidosis \\
3 & A-a O\textsubscript{2} Gradient & PaO\textsubscript{2}=68, PaCO\textsubscript{2}=54, FiO\textsubscript{2}=80\% & 434.9\,mmHg \\
\multicolumn{4}{c}{\textit{\ldots (7 additional steps omitted for brevity)}} \\
\bottomrule
\end{tabular}
}
\caption{Ground-truth workflow (subset) for the case.}
\label{tab:e2e-gt}
\end{table}

\textbf{GPT-5 Response.}
GPT-5 completed this task in 5 agent steps, using the \texttt{postgres\_executor} MCP server for SQL access and Python for computation.

Step~1 --- Schema exploration.
The model first called \texttt{list\_tables}, discovering 9 tables, then \texttt{describe\_tables} to inspect column names and types. No patient data was retrieved yet.

Steps~2--3 --- SQL data retrieval.
The model issued a multi-table SQL query joining \texttt{vital\_signs}, \texttt{laboratory\_result}, \texttt{examination\_report}, and \texttt{order\_inpatient}. The initial ABG sub-query used exact lowercase matches and returned only Lactate---the ABG items were stored as \texttt{pH (arterial)}, \texttt{PaCO2}, etc., which did not match. The model recognized the failure, adapted its query to use \texttt{LIKE '\%arterial\%' OR \ldots LIKE '\%blood gas\%'}, and successfully retrieved the full ABG panel. Ventilator parameters (PEEP, P\textsubscript{plat}, VT) were not found, as they are stored in free-text fields rather than structured columns.

Steps~4--5 --- Calculator selection and computation.
Using the retrieved data, the model computed the results shown in Table~\ref{tab:e2e-gpt5}.

\begin{table}[h]
\centering
\small
\resizebox{\columnwidth}{!}{
\setlength{\tabcolsep}{3pt}
\begin{tabular}{lcccc}
\toprule
\textbf{Calculator} & \textbf{Model Output} & \textbf{GT} & \textbf{Name} & \textbf{Result} \\
\midrule
P/F Ratio & 85\,mmHg & 85.0 & \ding{52} & \ding{52} \\
A-a O\textsubscript{2} Gradient & 434.9\,mmHg & 434.9 & \ding{52} & \ding{52} \\
SpO\textsubscript{2}/FiO\textsubscript{2} & \textit{SpO\textsubscript{2} not found} & 113.8 & \ding{52} & \ding{55} \\
Berlin Criteria & Partial & Positive & \ding{55} & \ding{55} \\
ETT Depth \& VT & No ETT depth & 22\,cm & \ding{55} & \ding{55} \\
Static Compliance & \textit{not found} & 24.4 & \ding{52} & \ding{55} \\
R/I Ratio & \textit{not attempted} & 0.39 & \ding{55} & \ding{55} \\
Murray Score & Partial & 3.8 & \ding{52} & \ding{55} \\
PRESET Score & \textit{not attempted} & 7 & \ding{55} & \ding{55} \\
ABG Analyzer & \textit{not performed} & Resp.\ Acid. & \ding{55} & \ding{55} \\
\midrule
\textbf{Total} & & & \textbf{5/10} & \textbf{2/10} \\
\bottomrule
\end{tabular}
}
\caption{GPT-5 results on the ARDS case. \ding{52} = correct match; \ding{55} = mismatch or missing.}
\label{tab:e2e-gpt5}
\end{table}

The model demonstrated accurate SQL schema exploration, adaptive query reformulation on failure, and correct arithmetic for the calculators it could complete. Its clinical narrative---recognizing severe ARDS, recommending low-tidal-volume ventilation, and flagging ECMO candidacy---was clinically appropriate despite incomplete computation.

\section{Implementation Details} \label{apd:details}

\subsection{Stage I Model Selection} \label{apd:stage1_validation}

Our data construction pipeline processes 1,970 initial calculators across 125 scenarios, involving substantial API calls. We selected models at each stage based on task characteristics and cost-performance trade-offs. Using \texttt{GPT-5} for all stages would increase the total cost by approximately $9\times$ and, being closed-source, would hinder community reproducibility. We therefore use well-established open-source models for Stage I and validate that their outputs are statistically equivalent to those of \texttt{GPT-5.2}.

\noindent \textbf{Calculator Deduplication with \texttt{GPT-OSS-120B}.} This step is a semantic similarity judgment task, requiring the identification of functionally identical calculators with different names. \texttt{GPT-OSS-120B} demonstrates strong medical semantic understanding at far lower cost than \texttt{GPT-5}. To validate this choice, we sampled 100 positive and 100 negative pairs and re-ran them with \texttt{GPT-5.2} under identical settings. As shown in Table~\ref{tab:stage1_dedup}, the two models reach 95\% overall agreement, with a Cohen's Kappa of 0.9 (where $\kappa > 0.8$ indicates near-equivalence). An MCC of 0.9045 and F1 of 0.9474 further corroborate the reliability of \texttt{GPT-OSS-120B} for this task and confirm its statistical equivalence with \texttt{GPT-5.2}.

\begin{table}[h]
\centering
\small
\begin{tabular}{lccc}
\toprule
 & \textbf{Positive} & \textbf{Negative} & \textbf{Overall} \\
\midrule
Consistent Rate & 90\% & 100\% & 95\% \\
Cohen's Kappa   & --   & --   & 0.9  \\
\bottomrule
\end{tabular}
\caption{Agreement between \texttt{GPT-OSS-120B} and \texttt{GPT-5.2} on calculator deduplication (100 positive and 100 negative pairs).}
\label{tab:stage1_dedup}
\end{table}

\noindent \textbf{Scenario Assignment with \texttt{DeepSeek-V3.1}.} Assigning calculators to the 125 clinical scenarios is essentially a classification-matching task. We randomly sampled 600 calculator--scenario pairs and compared \texttt{DeepSeek-V3.1} with \texttt{GPT-5.2}, obtaining a Cohen's Kappa of 0.855 ($>0.8$) and a Jaccard coefficient of 0.817, again confirming high agreement and statistical equivalence with \texttt{GPT-5.2}.

\subsection{Result Extraction Validation} \label{apd:extraction}

After each ReAct rollout, we use \texttt{DeepSeek-V3.1} to parse the agent's execution trajectory and extract the final calculator names, numerical results, and retrieved SQL fields into a standardized JSON format for automated evaluation. This step is a structured information extraction task that does not involve complex medical reasoning.

To validate this choice, we randomly sampled 20 tasks and re-ran the same extraction with \texttt{GPT-5.2} under identical prompts. Table~\ref{tab:extraction_agreement} reports per-dimension agreement rates,, and Cohen's Kappa, and Table~\ref{tab:extraction_correlation} reports cross-model correlations of per-task scores.

\begin{table}[h]
\centering
\small
\setlength{\tabcolsep}{4pt}
\begin{tabular}{lccc}
\toprule
\textbf{Dimension} & \textbf{Nums.} & \textbf{Agreement} & \textbf{$\kappa$} \\
\midrule
Calculator Name   & 127 & 93.7\% & 0.8423 \\
Calculator Result & 127 & 92.1\% & 0.8403 \\
SQL Field         & 652 & 82.5\% & 0.6454 \\
\midrule
\textbf{Overall}  & 906 & \textbf{85.4\%} & \textbf{0.7038} \\
\bottomrule
\end{tabular}
\caption{Per-dimension agreement between \texttt{DeepSeek-V3.1} and \texttt{GPT-5.2} on result extraction across 20 sampled tasks.}
\label{tab:extraction_agreement}
\end{table}

\begin{table}[h]
\centering
\small
\setlength{\tabcolsep}{6pt}
\begin{tabular}{lccc}
\toprule
\textbf{Dimension} & \textbf{Pearson $r$} & \textbf{Spearman $\rho$} & \textbf{MAD} \\
\midrule
Calculator Name   & 0.957 & 0.955 & 0.045 \\
Calculator Result & 0.919 & 0.864 & 0.061 \\
SQL Field         & 0.565 & 0.558 & 0.116 \\
\midrule
\textbf{Overall}  & \textbf{0.872} & \textbf{0.874} & \textbf{0.074} \\
\bottomrule
\end{tabular}
\caption{Cross-model correlation between per-task scores produced by \texttt{DeepSeek-V3.1} and \texttt{GPT-5.2}.}
\label{tab:extraction_correlation}
\end{table}

On the two dimensions most central to evaluation---calculator name recognition and calculator result matching---the two extractors reach near-perfect agreement ($\kappa > 0.84$). SQL field matching shows substantial agreement ($\kappa = 0.6454$), reflecting the higher granularity and occasional ambiguity of field-level judgments (e.g., synonymous column names). At the overall task-ranking level, Spearman $\rho = 0.874$ confirms that both extractors yield highly consistent model rankings. \texttt{DeepSeek-V3.1} is marginally stricter on average, but this bias is small and does not alter the ordering of evaluated models. These results support using \texttt{DeepSeek-V3.1} as a cost-effective and reproducible extractor in our evaluation pipeline.

\subsection{EHR Database} \label{apd:ehr}
To ensure the assessment tasks are based on real-world clinical scenarios, we constructed a dataset containing nine core tables based on MIMIC-IV v3.1. The structure of each table is shown in Tables~\ref{patient_info},~\ref{visit_in},~\ref{exam_report},~\ref{lab_result},~\ref{diag_record},~\ref{anethesia},~\ref{admission},~\ref{order} and~\ref{vital}.

\begin{table}[H]
\centering
\scriptsize
\caption{\texttt{patient\_information}}
\label{patient_info}
\begin{tabular}{>{\ttfamily}p{2cm} >{\ttfamily}p{1.5cm} >{\ttfamily}p{3cm}}
\toprule[0.4pt]
\textbf{Field Name} & \textbf{Type} & \textbf{Description} \\
\midrule[0.2pt]
medical\_\allowbreak institution\_code & TEXT & Medical institution code \\
patient\_id                & TEXT & Primary key, unique patient identifier \\
campus\_name               & TEXT & Campus name of the medical institution \\
patient\_name              & TEXT & Patient's full name \\
gender                     & TEXT & Patient's gender \\
birth\_date                & DATE & Patient's date of birth \\
visit\_card\_no            & TEXT & Internal visit card number \\
marriage\_status           & TEXT & Patient's marital status \\
ethnicity                  & TEXT & Patient's ethnic group \\
nationality                & TEXT & Patient's nationality \\
abo\_blood\_type           & TEXT & ABO blood group \\
rh\_blood\_type            & TEXT & Rh blood group \\
source\_system             & TEXT & Source system of the data \\
is\_valid                  & BOOLEAN & Validity indicator of the record \\
\bottomrule[0.4pt]
\end{tabular}
\end{table}

\begin{table}[H]
\centering
\scriptsize
\caption{\texttt{visit\_inpatient}}
\label{visit_in}
\begin{tabular}{>{\ttfamily}p{2cm} >{\ttfamily}p{1.5cm} >{\ttfamily}p{3cm}}
\toprule[0.4pt]
\textbf{Field Name} & \textbf{Type} & \textbf{Description} \\
\midrule[0.2pt]
medical\_\allowbreak institution\_code & TEXT & Medical institution code \\
patient\_id                & TEXT & Patient identifier (NOT NULL) \\
visit\_id                  & TEXT & Visit identifier (NOT NULL, PRIMARY KEY) \\
visit\_type                & TEXT & Visit type (reference code) (NOT NULL) \\
campus\_name               & TEXT & Campus name (NOT NULL) \\
inpatient\_no              & TEXT & Source inpatient number \\
visits\_num                & INTEGER & Visit count \\
patient\_name              & TEXT & Patient name (NOT NULL) \\
gender                     & TEXT & Gender (NOT NULL) \\
patient\_age               & INTEGER & Patient age \\
patient\_age\_unit         & TEXT & Age unit \\
admission\_time            & TIMESTAMP & Admission time \\
admission\_way             & TEXT & Admission way e.g. outpatient, emergency \\
admission\_dept\_name      & TEXT & Admission department name \\
admission\_dept\_code      & TEXT & Admission department source code \\
admission\_ward\_name      & TEXT & Admission ward name \\
admission\_ward\_code      & TEXT & Admission ward source code \\
major\_diagnosis           & TEXT & Main diagnosis \\
bed\_no                    & TEXT & Bed number \\
current\_dept\_name        & TEXT & Current department name \\
current\_dept\_code        & TEXT & Current department source code \\
current\_ward\_name        & TEXT & Current ward name \\
current\_ward\_code        & TEXT & Current ward source code \\
discharge\_dept\_name      & TEXT & Discharge department name \\
discharge\_dept\_code      & TEXT & Discharge department source code \\
discharge\_ward\_name      & TEXT & Discharge ward name \\
discharge\_ward\_code      & TEXT & Discharge ward source code \\
discharge\_time            & TIMESTAMP & Discharge time \\
create\_time               & TIMESTAMP & Record create time \\
update\_time               & TIMESTAMP & Record update time \\
is\_valid                  & BOOLEAN & Is valid \\
\bottomrule[0.4pt]
\end{tabular}
\end{table}

\begin{table}[H]
\centering
\scriptsize
\caption{\texttt{examination\_report}}
\label{exam_report}
\begin{tabular}{>{\ttfamily}p{2cm} >{\ttfamily}p{1.5cm} >{\ttfamily}p{3cm}}
\toprule[0.4pt]
\textbf{Field Name} & \textbf{Type} & \textbf{Description} \\
\midrule[0.2pt]
medical\_\allowbreak institution\_code & TEXT & Medical institution code \\
patient\_id                & TEXT & Patient identifier (NOT NULL) \\
visit\_id                  & TEXT & Visit identifier (NOT NULL) \\
report\_id                 & TEXT & Report identifier (NOT NULL, PRIMARY KEY) \\
campus\_name               & TEXT & Campus name (NOT NULL) \\
visit\_type                & TEXT & Visit type (NOT NULL) \\
patient\_name              & TEXT & Patient name (NOT NULL) \\
gender                     & TEXT & Gender (NOT NULL) \\
patient\_age               & INTEGER & Patient age \\
patient\_age\_unit         & TEXT & Age unit \\
bed\_no                    & TEXT & Bed number \\
clinic\_diag\_name         & TEXT & Clinical diagnosis \\
medical\_history           & TEXT & Medical history summary \\
exam\_type                 & TEXT & Examination type \\
exam\_item\_name           & TEXT & Exam item name \\
exam\_site                 & TEXT & Exam site \\
exam\_method               & TEXT & Exam method \\
exam\_tech                 & TEXT & Exam technique \\
report\_name               & TEXT & Report name \\
enhance\_scan\_flag        & TEXT & Enhancement flag \\
is\_anesthesia             & TEXT & Anesthesia flag \\
exam\_time                 & TIMESTAMP & Exam time \\
exam\_abnormal\_flag       & TEXT & Abnormal flag (1 normal, 2 abnormal, 3 uncertain) \\
exam\_findings             & TEXT & Findings \\
exam\_conclusion           & TEXT & Conclusion \\
radiology\_note            & TEXT & Radiology notes \\
intact\_exam\_report       & TEXT & Full report text \\
summary\_note              & TEXT & Summary description \\
comment                    & TEXT & Comment \\
review\_note               & TEXT & Review notes \\
application\_no            & TEXT & Application number \\
review\_time               & TIMESTAMP & Review time \\
report\_time               & TIMESTAMP & Report time \\
report\_state              & TEXT & Report state (0 void, 1 normal) \\
create\_time               & TIMESTAMP & Create time \\
update\_time               & TIMESTAMP & Update time \\
is\_valid                  & BOOLEAN & Is valid \\
\bottomrule[0.4pt]
\end{tabular}
\end{table}

\begin{table}[H]
\centering
\scriptsize
\caption{\texttt{laboratory\_result}}
\label{lab_result}
\begin{tabular}{>{\ttfamily}p{2cm} >{\ttfamily}p{1.5cm} >{\ttfamily}p{3cm}}
\toprule[0.4pt]
\textbf{Field Name} & \textbf{Type} & \textbf{Description} \\
\midrule[0.2pt]
medical\_\allowbreak institution\_code & TEXT & Medical institution code \\
patient\_id                & TEXT & Patient identifier (NOT NULL) \\
visit\_id                  & TEXT & Visit identifier (NOT NULL) \\
report\_id                 & TEXT & Report identifier (NOT NULL) \\
report\_item\_id           & TEXT & Report item identifier (NOT NULL, PRIMARY KEY) \\
campus\_name               & TEXT & Campus name (NOT NULL) \\
visit\_type                & TEXT & Visit type (NOT NULL) \\
test\_report\_name         & TEXT & Test package name \\
sort\_no                   & TEXT & Display order \\
test\_method               & TEXT & Test method \\
test\_item\_name           & TEXT & Test item name \\
test\_result               & TEXT & Test result \\
unit                       & TEXT & Result unit \\
sample\_name               & TEXT & Sample type \\
normal\_low                & FLOAT & Normal low \\
normal\_high               & FLOAT & Normal high \\
reference\_range           & TEXT & Reference range \\
abnormal\_flag             & TEXT & Abnormal flag \\
critical\_low              & FLOAT & Critical low \\
critical\_high             & FLOAT & Critical high \\
critical\_flag             & TEXT & Critical flag \\
absurd\_low                & FLOAT & Absurd low \\
absurd\_high               & FLOAT & Absurd high \\
report\_time               & TIMESTAMP & Report time \\
comment                    & TEXT & Comment \\
create\_time               & TIMESTAMP & Create time \\
update\_time               & TIMESTAMP & Update time \\
is\_valid                  & BOOLEAN & Is valid \\
\bottomrule[0.4pt]
\end{tabular}
\end{table}

\begin{table}[H]
\centering
\scriptsize
\caption{\texttt{diagnostic\_record}}
\label{diag_record}
\begin{tabular}{>{\ttfamily}p{2cm} >{\ttfamily}p{1.5cm} >{\ttfamily}p{3cm}}
\toprule[0.4pt]
\textbf{Field Name} & \textbf{Type} & \textbf{Description} \\
\midrule[0.2pt]
medical\_\allowbreak institution\_code & TEXT & Medical institution code \\
campus\_name               & TEXT & Campus name (NOT NULL) \\
patient\_id                & TEXT & Patient identifier (NOT NULL) \\
visit\_id                  & TEXT & Visit identifier (NOT NULL) \\
report\_id                 & TEXT & Document id \\
report\_item\_id           & TEXT & Diagnosis id (NOT NULL, PRIMARY KEY) \\
diagnosis\_\allowbreak
class\_name     & TEXT & Diagnosis class name \\
diagnosis\_code            & TEXT & Diagnosis source code \\
diagnosis\_seq             & TEXT & Diagnosis sequence \\
diagnosis\_name            & TEXT & Diagnosis name \\
main\_flag                 & TEXT & Main diagnosis flag \\
diagnosis\_source          & TEXT & Source of diagnosis \\
diagnosis\_time            & TIMESTAMP & Diagnosis time \\
create\_time               & TIMESTAMP & Create time \\
update\_time               & TIMESTAMP & Update time \\
is\_valid                  & BOOLEAN & Is valid \\
\bottomrule[0.4pt]
\end{tabular}
\end{table}

\begin{table}[H]
\centering
\scriptsize
\caption{\texttt{anethesia\_record}}
\label{anethesia}
\begin{tabular}{>{\ttfamily}p{2cm} >{\ttfamily}p{1.5cm} >{\ttfamily}p{3cm}}
\toprule[0.4pt]
\textbf{Field Name} & \textbf{Type} & \textbf{Description} \\
\midrule[0.2pt]
medical\_\allowbreak institution\_code & TEXT & Medical institution code \\
campus\_name               & TEXT & Campus name (NOT NULL) \\
patient\_id                & TEXT & Patient identifier (NOT NULL) \\
visit\_id                  & TEXT & Visit identifier (NOT NULL) \\
report\_id                 & TEXT & Anesthesia record id (NOT NULL, PRIMARY KEY) \\
visit\_type                & TEXT & Visit type (NOT NULL) \\
height                     & FLOAT & Height in cm \\
weight                     & FLOAT & Weight in kg \\
urine\_volume              & FLOAT & Urine volume in ml \\
blood\_lossed              & FLOAT & Blood loss in ml \\
asa\_class\_code           & TEXT & ASA class \\
surgical\_position         & TEXT & Surgical position \\
whether\_fast              & TEXT & Fasting flag \\
preoperative\_\allowbreak diagnosis    & TEXT & Preoperative diagnosis \\
propose\_\allowbreak surgery \_name     & TEXT & Proposed surgery name \\
intraoperative\_\allowbreak surgical\_name & TEXT & Intraoperative surgery name \\
breath\_type               & TEXT & Breath type \\
tracheal\_\allowbreak intubation\_type & TEXT & Intubation type \\
drug\_before\allowbreak \_anesthesia   & TEXT & Pre-anesthesia drugs \\
anesthesia\allowbreak \_start\_time    & TIMESTAMP & Start time \\
anesthesia\_end\_time      & TIMESTAMP & End time \\
anesthesia\_type           & TEXT & Anesthesia type \\
anesthesia\allowbreak \_real\_duration & FLOAT & Real duration in hours \\
go\_after\_operation       & TEXT & Post-op disposition \\
create\_time               & TIMESTAMP & Create time \\
update\_time               & TIMESTAMP & Update time \\
is\_valid                  & BOOLEAN & Is valid \\
\bottomrule[0.4pt]
\end{tabular}
\end{table}

\begin{table}[H]
\centering
\scriptsize
\caption{\texttt{admission\_record}}
\label{admission}
\begin{tabular}{>{\ttfamily}p{2cm} >{\ttfamily}p{1.5cm} >{\ttfamily}p{3cm}}
\toprule[0.4pt]
\textbf{Field Name} & \textbf{Type} & \textbf{Description} \\
\midrule[0.2pt]
medical\_\allowbreak institution\_code & TEXT & Medical institution code \\
patient\_id                & TEXT & Patient identifier (NOT NULL) \\
visit\_id                  & TEXT & Visit identifier (NOT NULL) \\
report\_id                 & TEXT & Document id (NOT NULL, PRIMARY KEY) \\
inpatient\_no              & TEXT & Inpatient number (NOT NULL) \\
campus\_name               & TEXT & Campus name (NOT NULL) \\
patient\_name              & TEXT & Patient name (NOT NULL) \\
gender                     & TEXT & Gender (NOT NULL) \\
admission\_time            & TIMESTAMP & Admission time \\
admission\_dept\_name      & TEXT & Admission department \\
admission\_dept\_code      & TEXT & Admission department code \\
admission\_ward\_name      & TEXT & Admission ward \\
admission\_ward\_code      & TEXT & Admission ward code \\
record\_title              & TEXT & Record title \\
document\_content          & TEXT & Document content \\
chief\_complaints          & TEXT & Chief complaints \\
present\_illness           & TEXT & Present illness \\
past\_medical\_history     & TEXT & Past medical history \\
personal\_history          & TEXT & Personal history \\
marital\_\allowbreak obstetrical\_history & TEXT & Marital/obstetrical history \\
family\_history            & TEXT & Family history \\
physical\allowbreak \_examination      & TEXT & Physical exam \\
accessory\allowbreak \_examination     & TEXT & Auxiliary exams \\
special\_examination       & TEXT & Specialty exams \\
intact\_physical\allowbreak \_examination & TEXT & Combined physical exam text \\
admission\_diagnosis       & TEXT & Admission diagnosis \\
initial\_diagnosis         & TEXT & Initial diagnosis \\
modified\_diagnosis        & TEXT & Modified diagnosis \\
supplementary\allowbreak \_diagnosis   & TEXT & Supplementary diagnosis \\
discharge\_diagnosis       & TEXT & Discharge diagnosis \\
record\_time               & TIMESTAMP & Record time \\
record\_state\_name        & TEXT & Record state \\
creator\_name              & TEXT & Creator name \\
create\_time               & TIMESTAMP & Create time \\
update\_time               & TIMESTAMP & Update time \\
is\_valid                  & BOOLEAN & Is valid \\
\bottomrule[0.4pt]
\end{tabular}
\end{table}

\begin{table}[H]
\centering
\scriptsize
\caption{\texttt{order\_inpatient}}
\label{order}
\begin{tabular}{>{\ttfamily}p{2cm} >{\ttfamily}p{1.5cm} >{\ttfamily}p{3cm}}
\toprule[0.4pt]
\textbf{Field Name} & \textbf{Type} & \textbf{Description} \\
\midrule[0.2pt]
medical\_\allowbreak institution\_code & TEXT & Medical institution code \\
patient\_id                & TEXT & Patient identifier (NOT NULL) \\
visit\_id                  & TEXT & Visit identifier (NOT NULL) \\
report\_id                 & TEXT & Order id (NOT NULL, PRIMARY KEY) \\
inpatient\_no              & TEXT & Inpatient number \\
campus\_name               & TEXT & Campus name (NOT NULL) \\
patient\_name              & TEXT & Patient name (NOT NULL) \\
group\_no                  & TEXT & Group number \\
group\_seq                 & TEXT & Group sequence \\
order\_class               & TEXT & Order class \\
order\_type                & TEXT & Order type \\
drug\_flag                 & TEXT & Is drug \\
order\_name                & TEXT & Order name \\
specification             & TEXT & Specification \\
dose\_form                 & TEXT & Dose form \\
unit\_price                & FLOAT & Unit price \\
quantity                  & FLOAT & Quantity \\
quantity\_unit             & TEXT & Quantity unit \\
once\_dose                 & FLOAT & Dose per time \\
dose\_unit                 & TEXT & Dose unit \\
frequency                 & TEXT & Frequency \\
special\_execution\allowbreak \_time   & TEXT & Special execution time \\
special\_execution\allowbreak \_dose   & TEXT & Special execution dose \\
usage                      & TEXT & Usage \\
procedure\_name            & TEXT & Procedure name \\
herbal\_payments           & INTEGER & Herbal payments \\
execution\_dept            & TEXT & Execution department \\
skin\_test                 & TEXT & Skin test flag \\
urgent\_flag               & TEXT & Urgent flag (0 normal, 1 urgent) \\
is\_discharge\allowbreak \_medicine    & TEXT & Discharge medicine flag \\
order\_advice              & TEXT & Order advice \\
order\_begin\_time         & TIMESTAMP & Order begin time \\
order\_end\_time           & TIMESTAMP & Order end time \\
dept\_name                 & TEXT & Department name \\
dept\_code                 & TEXT & Department code \\
ward\_name                 & TEXT & Ward name \\
ward\_code                 & TEXT & Ward code \\
submit\_dept\_name         & TEXT & Submit department name \\
submit\_dept\_code         & TEXT & Submit department code \\
submit\_time               & TIMESTAMP & Submit time \\
body\_site\_name           & TEXT & Body site name \\
sample\_name               & TEXT & Sample name \\
confirm\_time              & TIMESTAMP & Confirm time \\
cancel\_time               & TIMESTAMP & Cancel time \\
execution\_time            & TIMESTAMP & Execution time \\
order\_state\_name         & TEXT & Order state name \\
create\_time               & TIMESTAMP & Create time \\
update\_time               & TIMESTAMP & Update time \\
is\_valid                  & BOOLEAN & Is valid \\
\bottomrule[0.4pt]
\end{tabular}
\end{table}

\begin{table}[H]
\centering
\scriptsize
\caption{\texttt{vital\_signs}}
\label{vital}
\begin{tabular}{>{\ttfamily}p{2cm} >{\ttfamily}p{1.5cm} >{\ttfamily}p{3cm}}
\toprule[0.4pt]
\textbf{Field Name} & \textbf{Type} & \textbf{Description} \\
\midrule[0.2pt]
medical\_\allowbreak institution\_code & TEXT & Medical institution code \\
patient\_id                & TEXT & Patient identifier (NOT NULL) \\
visit\_id                  & TEXT & Visit identifier (NOT NULL) \\
vs\_id                     & TEXT & Vital sign id (NOT NULL, PRIMARY KEY) \\
campus\_name               & TEXT & Campus name (NOT NULL) \\
patient\_name              & TEXT & Patient name (NOT NULL) \\
dept\_name                 & TEXT & Department name \\
dept\_code                 & TEXT & Department code \\
ward\_name                 & TEXT & Ward name \\
ward\_code                 & TEXT & Ward code \\
bed\_no                    & TEXT & Bed number \\
plan\_time                 & TIMESTAMP & Planned time \\
measure\_time              & TIMESTAMP & Measured time \\
breathe                    & TEXT & Respiration (per min) \\
pulse                      & TEXT & Pulse (per min) \\
heart\_rate                & TEXT & Heart rate (per min) \\
temperature               & TEXT & Temperature (°C) \\
systolic\_pressure         & TEXT & Systolic pressure (mmHg) \\
diastolic\_pressure        & TEXT & Diastolic pressure (mmHg) \\
height                     & TEXT & Height (cm) \\
weight                     & TEXT & Weight (kg) \\
defecate\_frequency        & TEXT & Defecation frequency (per day) \\
urine\_volume              & TEXT & Urine volume (ml) \\
sputum\_volume             & TEXT & Sputum volume (ml) \\
drainage\_volume           & TEXT & Drainage volume (ml) \\
emesis\_volume             & TEXT & Emesis volume (ml) \\
output\_total\_volume      & TEXT & Total output volume (ml) \\
income\_volume             & TEXT & Intake volume (ml) \\
postoperative\_days        & INTEGER & Postoperative days \\
after\_delivery\_days      & INTEGER & Days after delivery \\
days\_in\_hospital         & INTEGER & Days in hospital \\
create\_time               & TIMESTAMP & Create time \\
update\_time               & TIMESTAMP & Update time \\
is\_valid                  & BOOLEAN & Is valid \\
\bottomrule[0.4pt]
\end{tabular}
\end{table}

\subsection{Task Construction} \label{apd:task_construction}
We use GPT-5 to generate evaluation tasks through a multi-stage process, and we designed a set of prompt templates that are applied at different steps.

\textbf{Task Generation}
\begin{lstlisting}[breaklines=true, breakatwhitespace=true, basicstyle=\small\ttfamily, columns=fullflexible]
Purpose: Generate ONE clinical workflow task within a coherent clinical scenario that requires the use of multiple medical calculators.

Context:
- Core Clinical Scenario: {major_scenario} _ {sub_category}
- Available calculators (exact names and short descriptions):
{calculators}

Requirements (CRITICAL):
1) Selection rules:
   - Prefer selecting calculators from the provided list. Use calculators outside the provided list **only if** the provided calculators are too few and clearly cannot cover a coherent, complete clinical workflow for the scenario. If you include any calculators not in the provided list, explicitly name them in "required_calculators" **and** provide a clear, one-line clinical justification for each in "scenario_analysis.workflow_logic". Do NOT invent or fabricate calculator names--only use established, widely-known calculators when adding external ones.
   - Aim to select **4 to 10** calculators when possible to create a comprehensive workflow. If you select 8-10, include a short justification in `scenario_analysis.workflow_logic` explaining why a larger set was needed for clinical completeness. If fewer than 4 relevant calculators exist in the provided list, you are **not required** to force the count up to 4; instead, produce a clinically coherent workflow using the appropriate number of calculators and explain the rationale in `scenario_analysis.workflow_logic`.
   - Choose calculators that collectively support a comprehensive clinical assessment or planning process for the given scenario. The primary criterion is **clinical appropriateness and workflow coherence**: prioritize selections that make the sequence logical and useful for decision-making, even if that means selecting fewer calculators than the target range.

2) Workflow Organization: 
   - Organize the selected calculators into a logical sequence that reflects a typical clinical workflow for this scenario. 
   - The connection between calculators should be based on clinical reasoning, ensuring a natural progression from one step to the next. 
   - Multiple calculators' outputs (at least one) must serve as a required input or deterministic decision trigger for the subsequent calculator(s).
   - Include multiple conditional branches(if/else) (at least one) based on an intermediate result. This will demonstrate how the outcome of one step informs or leads to the next.

3) Content constraints:
   - DO NOT include any patient-specific numeric values, lab names, demographics, or external resource references (no URLs, files, DBs, APIs).
   - Describe a unified clinical scenario and then specify the sequence of calculator uses within it.
   - Focus on the purpose of each calculator in the context of the overall clinical workflow.

4) Output format (MUST be the only output; no extra text):
Return exactly one valid JSON object that matches the schema below. If you cannot produce the required JSON, return this exact error object instead: {"error":"unable_to_comply"}.

Schema:
{
  "task_id": "medical_workflow_[UNIQUE_ID_HERE]",
  "task_title": "short title reflecting the core scenario",
  "task_description": "Full natural-language task: Start with a cohesive clinical scenario description for '{major_scenario} _ {sub_category}'. Then, detail the step-by-step process of using the selected calculators in a logical order to address this scenario. Describe the clinical rationale for the sequence.",
  "required_calculators": ["Exact Name of Selected Calc A", "Exact Name of Selected Calc B", ...],
  "scenario_analysis": {
    "scenario_rationale": "Explain why this set of calculators provides comprehensive coverage for the core clinical scenario.",
    "workflow_logic": "Describe the clinical logic behind the order of steps (assessment, decision points, planning stages).",
    "clinical_goal": "State the overall clinical objective achieved by completing this workflow."
  }
}

Strict rules to follow:
- Only use calculator names exactly as provided in the {calculators} list.
- The "required_calculators" array must contain only the selected calculators, ordered logically.
- Output MUST be valid JSON parseable by a strict JSON parser (no comments, no trailing commas, use double quotes).
- Output ONLY the JSON object and nothing else.
\end{lstlisting}

\textbf{Task Fuzzy}

\begin{lstlisting}[breaklines=true, breakatwhitespace=true, basicstyle=\small\ttfamily, columns=fullflexible]
Purpose: Convert detailed clinical calculation tasks into natural, conversational medical requests that still implicitly but completely encompasses each step of the original task flow and the required intermediate deliverables. 
This requires the executor to solve the problem step by step according to the complete workflow in order to arrive at a justifiable conclusion.
Convert this detailed clinical calculation task into a NATURAL, CONVERSATIONAL MEDICAL REQUEST that truly tests the agent's clinical reasoning ability.

Original detailed task: {detailed_task}

Available tools: {len_tools} clinical calculators (but DON'T mention them in the fuzzy version)

CRITICAL: CREATE A GENUINELY NATURAL CLINICAL REQUEST

ABSOLUTELY FORBIDDEN:
- ANY structured language that looks like a clinical protocol or algorithm
- Phrases like "I need to calculate", "Please compute", "Determine the value of"
- ANY specific calculator names or technical tool references
- Formal medical terminology used in isolation

INSTEAD, CREATE A NATURAL CLINICAL CONVERSATION:
- Start with patient context or clinical uncertainty: "I'm seeing a patient who...", "We have a clinical situation where..."
- Use conversational medical openers: "I'm trying to figure out the best approach for...", "Been reviewing a case where...", "Got a patient where we're not sure about..."
- Include clinical uncertainty: "not certain if", "wondering whether", "could be", "might need to consider"
- Add clinical context: "for my patient management", "in our clinic we're discussing", "I'm reviewing this case for"
- Express the clinical need through a patient story or scenario, not a calculation checklist

REQUIRED: STRUCTURED REASONING, BUT DISGUISED AS CLINICAL THINKING
You MUST encode every step from the original task, but expressed as:
- Clinical doubts
- Sequential concerns
- "If this is the case, then..." style reasoning
- Management-oriented thinking
Think in terms of clinical dependency, not math steps.

STRICT REQUIREMENTS: The output must cover every step within task_description (even if there are multiple sub-steps in the original text, they must be presented as clinical task points in the fuzzy request), so that the executor cannot skip intermediate results and directly draw conclusions. But never tell them directly which calculator or tool to use.

HIDE THE CALCULATION STRUCTURE COMPLETELY:

Don't say: "Calculate BMI, then BSA, then adjust chemotherapy dose based on renal function"
Say instead: "I am managing a cancer patient who requires chemotherapy, but I want to ensure precise dosing. I need to consider the patient's level of obesity and physical condition, and adjust the chemotherapy dose based on renal function."

Don't say: "Please calculate the patient's CHA2DS2-VASc score"
Say instead: "Given the patient's age, comorbidities, and vascular history, systematic assessment of his thromboembolic risk is required."


PRESERVE CLINICAL CONTEXT NATURALLY:
- Embed patient factors conversationally: "middle-aged", "somewhat overweight", "kidney function isn't perfect"
- Use approximate clinical language: "roughly in this range", "about this level", "somewhere around"
- Keep exact clinical concepts when necessary but phrase naturally

MAKE IT SOUND LIKE A REAL CLINICAL DISCUSSION:
- Use natural medical dialogue: "What's your clinical thinking on this?", "How would you approach this case?"
- Include realistic clinical hesitation: "I'm torn between...", "Part of me thinks X, but then there's Y to consider"
- Show appropriate clinical concern: "really want to avoid complications", "been worrying about this aspect"
- Ask for clinical reasoning naturally: "What's your take?", "How would you reason through this?", "Am I missing anything here?"

CRITICAL: End naturally with clinical evidence requirements:
Instead of: "Please provide evidence-based calculations"
Say: "I really need proper calculations on this - can't make this decision based on gut feeling alone. Whatever approach you suggest, make sure it's backed by solid calculations, okay? This needs to hold up to scrutiny."

ALWAYS USE clinical timeframes naturally:
- "recent labs show", "over the past few visits", "looking ahead to their next appointment"
- NOT specific dates like "January 15th" or "in 2024"

Return ONLY the natural, conversational fuzzy description - make it sound like a real clinician discussing a case, not a robot executing calculations.
\end{lstlisting}

\textbf{Remove Patient Data}
\begin{lstlisting}[breaklines=true, breakatwhitespace=true, basicstyle=\small\ttfamily, columns=fullflexible]
Please process the following medical question text by removing all direct personal information and specific data about the patient, while maintaining the core content of the question and the completeness of the medical discussion.

Processing requirements:
1. Remove all information that could identify the patient
2. Remove specific test values, dates, ages, and other numerical data
3. Convert specific medical history descriptions to general descriptions
4. Preserve the essence of the medical question and the logic of clinical decision-making
5. Maintain the original tone and purpose of the inquiry

Example input:
I'm reviewing a case for a patient with cirrhosis who we just found has a liver tumor, an HCC. We're considering surgery to remove it, but I'm really hesitant. His liver function isn't great to begin with, and I'm worried a big operation could push him over the edge into full decompensation. I need to get a solid handle on whether he can even tolerate a hepatectomy. Part of me thinks we should just go for it, but then I keep worrying about his heart--he's got some risk factors there too, and major surgery is a huge stressor. And on top of that, I'm concerned about his lungs post-op; he's a former smoker. We absolutely need to avoid a situation where we cause more problems than we solve. I'm trying to figure out the best path forward. Is resection even a safe option for him, or are we looking at a transplant evaluation instead? I really need some evidence-based risk stratification here--something more than just my clinical gut feeling. Can you walk me through how you'd assess his overall surgical risk? Whatever approach you take, make sure it's backed by proper calculations. This decision has to be rock-solid.

Expected output:
I'm reviewing a case for a patient with cirrhosis who has a liver tumor (HCC). We're considering surgery, but I'm hesitant. The patient's liver function is compromised, and I'm worried a major operation could lead to decompensation. I need to determine surgical tolerance for hepatectomy. I'm torn between proceeding with surgery and being cautious. The patient has cardiac risk factors, and major surgery poses significant stress. Additionally, there's a history of smoking, raising concerns about post-operative pulmonary complications. We need to avoid causing more harm than benefit. I'm trying to determine the optimal approach. Is resection a safe option, or should we consider transplant evaluation? I need evidence-based risk stratification beyond clinical intuition. Can you guide me through a comprehensive surgical risk assessment? The approach should be supported by proper calculations for a well-founded decision.

Please process the following text:
{INSERT_TEXT_HERE}
\end{lstlisting}

\textbf{Create Patient Data}
\begin{lstlisting}[breaklines=true, breakatwhitespace=true, basicstyle=\small\ttfamily, columns=fullflexible]
Your task is to generate realistic, consistent, and comprehensive patient data required to execute the complete clinical calculator task described below. 
The data must align with the clinical scenario and information in the task. 
You must output clear numerical values or descriptions, not ambiguous terms. 
If relevant data does not exist, you may design it appropriately.

Context:
- Clinical Task Description: {task_description}
- Fuzzy Description: {fuzzy_description}
- Available Medical Calculators:
{tool_descriptions}

Output the necessary inputs for calculating ALL calculators mentioned in the task_description, in sequential order. 
Ensure the generated data is consistent with all contextual clues and falls within clinically plausible ranges.

Output ONLY the JSON object, with no explanatory text.
Output Format:
{
  "metadata": {
    "scenario_reference": "brief text summarizing how data matches the scenario",
    "note": "de-identified, synthetic or derived data -- not real PII"
  },
  "calculators": [
    {
      "order": 1,
      "name": "Exact Calculator Name (must match provided list)",
      "inputs": [
        { "field": "field_name", "value": "note or value (with unit)" }
        // ... (Add more inputs for this calculator)
      ],
      "notes": "optional short note about assumptions or mapping from scenario"
    }
    // ... (Follow the calculators order exactly. Add more objects for all calculators)
  ]
}
\end{lstlisting}

\textbf{Create Case Report}
\begin{lstlisting}[breaklines=true, breakatwhitespace=true, basicstyle=\small\ttfamily, columns=fullflexible]
You are a senior attending physician. The task is to generate a concise and realistic clinical report based on the provided partial patient data. You should include the chief complaint, physical examination findings, and diagnostic reports.

CRITICAL REQUIREMENTS:

DO NOT mention any calculator names, calculator results, or scoring systems.
Present all values as if they were obtained from real hospital systems (EHR notes, lab results, imaging reports, vital signs, etc.).
Focus only on the available data, and supplement with reasonable and medically accurate details where appropriate to complete the clinical picture.
The report should be concise, including only the most essential sections, and limited to the data that has been provided up until this point.
Maintain clinical coherence, using realistic medical language and making sure that all data is consistent with standard medical practice.

INPUT INFORMATION:
Clinical Scenario: {scenario}
Clinical Task: {task_description}
Patient Data: {patient_data}
Reference Cases: {reference_cases}

REPORT REQUIREMENTS:
Produce a full clinical case report based on the current available data. This should include only the following sections:
- PATIENT DEMOGRAPHICS(Name, sex, age, relevant background information)
- CHIEF COMPLAINT(The main issue or concern that led the patient to seek care)
- PHYSICAL EXAMINATION
- Vital signs, system-by-system examination findings, as available at the moment.
- DIAGNOSTIC REPORTS
- Relevant laboratory test results, imaging findings, or other diagnostics available at this time.

ADDITIONAL GUIDELINES:
- Ensure the language used is appropriate for a medical report, with clear headings for each section.
- The report should be concise and medically plausible, integrating the data provided with realistic values and observations.
- No unnecessary sections such as treatment plans or discharge instructions should be included.
- Ensure no contradictions with the provided data or clinical scenario.

Now, generate the clinical report based on the provided data.
\end{lstlisting}

\textbf{Data Reformat}

\begin{lstlisting}[breaklines=true, breakatwhitespace=true, basicstyle=\small\ttfamily, columns=fullflexible]
Your task is to reorganize patient data from a calculator-based structure into a structured format that fits nine database tables. 
You need to map the input data and the clinical context to the appropriate fields in each table, ensuring no data is lost and resolving any conflicts by selecting the most appropriate value.

***IMPORTANT CRITICAL REQUIREMENT:***
**The values MUST mimic those on a normal hospital laboratory report or clinical documentation.**
**You MUST NOT directly include any calculator names, calculator results, or any references to calculators in the output.**
**Transform all data into standard hospital record format as if it came from actual clinical systems (EHR, lab reports, nursing notes, etc.).**

TASK_ID: {task_id}
SCENARIO: {scenario}
INPUT DATA: {data}
CLINICAL CONTEXT: {context}

DATABASE TABLES STRUCTURE:
You have nine tables to populate.
SCHEMA CODE FOR DATABASE: {schema}
**Conflict and Selection Principles**:
   - If an input can be mapped to multiple tables, fill it in all relevant tables while maintaining data consistency across them.
   - For synonymous field names (e.g., `sbp`/`systolic_bp`/`systolic`), uniformly choose the **most detailed and explicit name** (e.g., `systolic_blood_pressure` if available, otherwise `systolic_bp`).

**Rules For Primary Keys Generation**:
1) Derive patient_id from the provided task_id using this exact logic:
   - Use the suffix of task_id to create the patient_id, set patient_id = "P" + suffix.
     Example: task_id = "medical_workflow_001_644ebd11" -> suffix = "001_644ebd11" -> patient_id = "P001_644ebd11"
   - Otherwise, set patient_id = "P_task_id". Example: task_id = "task123" -> patient_id = "P_task123".
2) Set visit_id = patient_id + "_V01" for every visit_inpatient row. (Example: patient_id = "P001_644ebd11" -> visit_id = "P001_644ebd11_V01")
3) For every non-empty table (except patient_information and visit_inpatient already handled), ensure the table-specific primary-key exists:
   - Use patient_id as prefix and append "(table_name)_n" where n is the 1-based row index in that table (order-preserving).
   - Example for laboratory_result: first row -> report_item_id = ""patient_id"+ "_RI_1", second row -> ""patient_id"+"_RI_2". If the primary key is report_id, use patient_id + "_R_n" instead.
   - If an existing primary-key already **starts with** patient_id + "_" keep it unchanged.
4) Do not modify any other fields. Return only the modified `reformat_data` JSON object.

OUTPUT FORMAT: The output should be a JSON object with nine arrays, each representing a table in the database. Each array should contain objects with the appropriate fields and values.
The values should mimic those on a normal hospital laboratory report and must not directly include the results from any calculators.
Strictly follow the format, output only the JSON, without any explanations or additional content.
```json
{
  "patient_information": [ { ... } ],
  "visit_inpatient": [ { ... } ],
  "vital_signs": [ { ... } ],
  "examination_report": [ { ... } ],
  "laboratory_result": [ { ... } ],
  "admission_record": [ { ... } ],
  "diagnostic_record": [ { ... } ],
  "order_inpatient": [ { ... } ],
  "anethesia_record": [ { ... } ]
}
\end{lstlisting}

\subsection{Recruitment} \label{apd:recruitment}
This study was conducted under continuous medical supervision. Licensed physicians were recruited to provide professional proofreading and participate in experimental procedures, with compensation at \$40 per hour.

\end{document}